\documentclass{article}

\usepackage{microtype}
\usepackage{graphicx}
\usepackage{subfigure}
\usepackage{booktabs} 
\usepackage{mdframed}
\usepackage{tcolorbox}

\usepackage{adjustbox}
\usepackage{tabularx}

\usepackage{hyperref}

\usepackage[accepted]{icml2025}

\usepackage{amsmath}
\usepackage{amssymb}
\usepackage{mathtools}
\usepackage{amsthm}

\usepackage{booktabs} 
\usepackage{multirow}
\usepackage{colortbl}
\usepackage{makecell}

\usepackage{algorithm}
\usepackage{algorithmic}

\usepackage{adjustbox}

\usepackage[capitalize,noabbrev]{cleveref}

\theoremstyle{plain}

\theoremstyle{definition}

\theoremstyle{remark}

\usepackage[textsize=tiny]{todonotes}

\begin{document}

\twocolumn[
\icmltitle{CL3DOR: Contrastive Learning for 3D Large Multimodal Models via Odds Ratio on High-Resolution Point Clouds}



\icmlsetsymbol{equal}{*}

\begin{icmlauthorlist}

\icmlauthor{Keonwoo Kim}{equal}
\icmlauthor{Yeongjae Cho}{equal}
\icmlauthor{Taebaek Hwang}{equal}
\icmlauthor{Minsoo Jo}{}
\icmlauthor{Sangdo Han}{}
\end{icmlauthorlist}




\vskip 0.3in
]





\renewcommand{\thefootnote}{\fnsymbol{footnote}}
\setcounter{footnote}{1} 
\footnotetext[1]{denotes equal contributions.}

\begin{abstract}
Recent research has demonstrated that Large Language Models (LLMs) are not limited to text-only tasks but can also function as multimodal models across various modalities, including audio, images, and videos. In particular, research on 3D Large Multimodal Models (3D LMMs) is making notable strides, driven by the potential of processing higher-dimensional data like point clouds. However, upon closer examination, we find that the visual and textual content within each sample of existing training datasets lacks both high informational granularity and clarity, which serve as a bottleneck for precise cross-modal understanding. To address these issues, we propose \textbf{CL3DOR}, \textbf{C}ontrastive \textbf{L}earning for \textbf{3D} large multimodal models via \textbf{O}dds ratio on high-\textbf{R}esolution point clouds, designed to ensure greater specificity and clarity in both visual and textual content. Specifically, we increase the density of point clouds per object and construct informative hard negative responses in the training dataset to penalize unwanted responses. To leverage hard negative responses, we incorporate the odds ratio as an auxiliary term for contrastive learning into the conventional language modeling loss. CL3DOR achieves state-of-the-art performance in 3D scene understanding and reasoning benchmarks. Additionally, we demonstrate the effectiveness of CL3DOR's key components through extensive experiments.
\end{abstract}

\section{Introduction}

Humans primarily understand and reason about the world based on vision and language. To emulate the human approach, recent research has focused on Large Multimodal Models (LMMs), particularly 2D LMMs~\cite{bai2023qwen, liu2024visual, jian2024bootstrapping, reid2024gemini}, which inject 2D vision features into the language model space to create general-purpose assistants. However, relying solely on 2D visual information poses significant limitations in accurately interpreting the 3D world. Since reasoning and understanding based on 3D data offer a higher-dimensional perspective, recent research has increasingly focused on 3D LMMs~\cite{xu2023pointllm, zhou2023uni3d, huang2023chat3dv2, huang2023embodied, qi2024gpt4point} that integrate 3D vision features into the language model space to improve the comprehension of 3D data and language instructions.

\begin{figure}[t]
\setlength{\abovecaptionskip}{0.2cm}
\setlength{\belowcaptionskip}{-0.2cm}
\centering
\includegraphics[width=8.2cm]{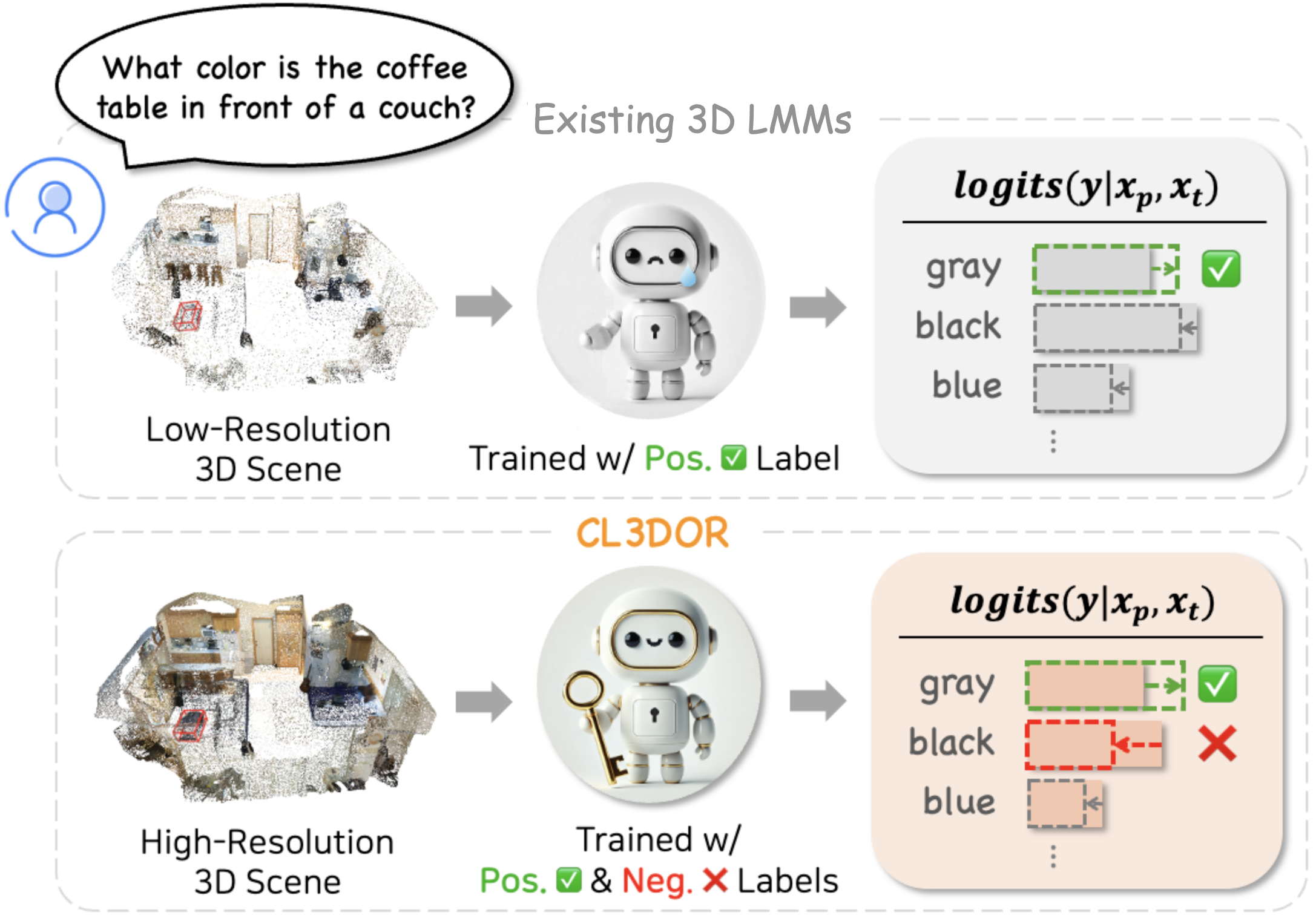}
\caption{Comparison of training methods for 3D LMMs. The upper side shows existing 3D LMMs trained with sparse (low-resolution) point clouds and positive labels, aiming to maximize logits for correct responses. The lower side illustrates the CL3DOR, which uses dense (high-resolution) point cloud input and incorporates both positive and negative labels, employing a contrastive learning approach to explicitly leverage negative response.}
\label{fig:introduction}
\end{figure}

\begin{figure*}[!ht]
\setlength{\abovecaptionskip}{0.2cm}
\centering
\includegraphics[width=17.5cm]{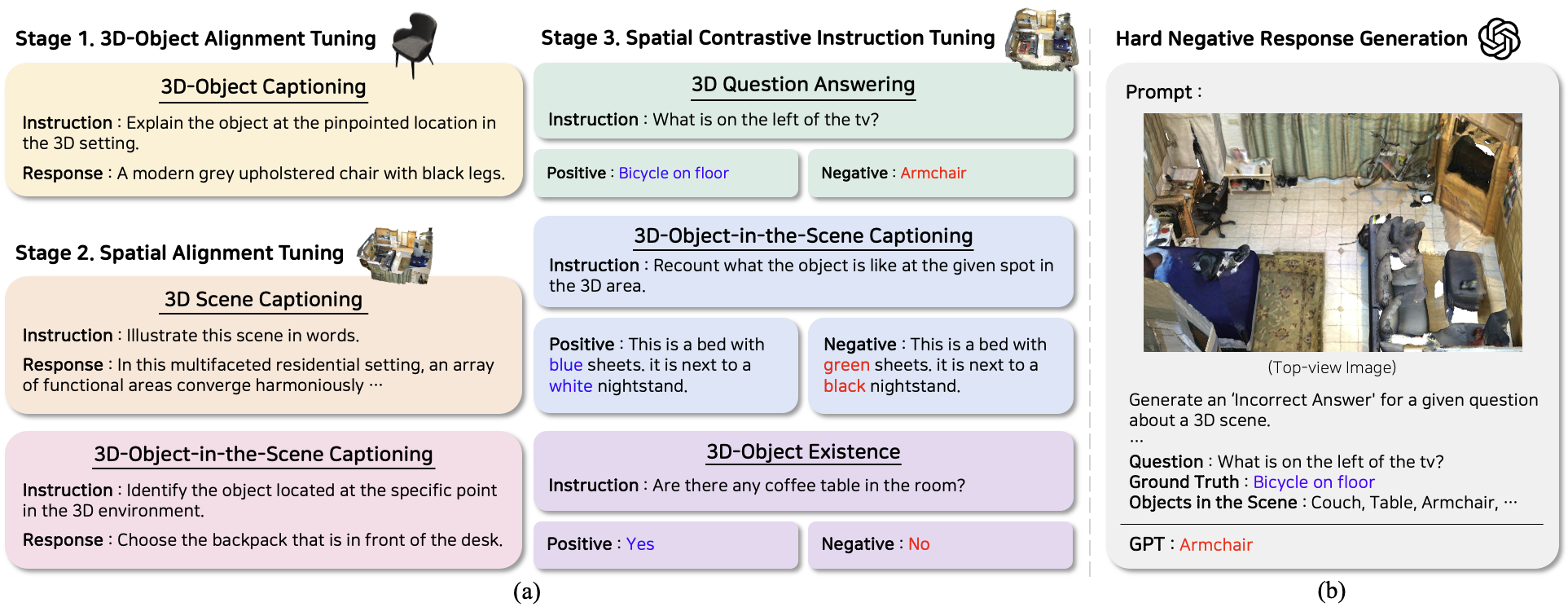}
\caption{(a) Examples of the three-stage training datasets used in CL3DOR. Notably, only Stage 3 includes two types of responses following an instruction for contrastive learning. (b) The process of generating hard negative responses for the 3D question answering task. We use GPT-4o to create plausible hard negatives, referencing a top-view image and scene objects}
\label{fig:datasets}
\end{figure*}

Recently, most 3D LMMs have been developed with a point encoder for processing point clouds, a pre-trained large language model (LLM)~\cite{jiang2023mistral, touvron2023llama, team2024gemma} for handling instructions, and projection layers that bridge the feature spaces of the point encoder and the LLM. In 3D LMM-related tasks, ensuring high performance is crucial, as performance degradation can lead to significant risks, including physical accidents when such models are deployed in real-world applications involving devices that exert physical force, such as robots~\cite{patil2023advances, sharkawy2022human}. However, we find that existing 3D LMMs often fall short in spatial understanding due to the quality of the training data. In line with recent studies~\cite{chen2024your, zhou2024lima}, we define high-quality training data as follows: \textbf{\textit{high informational granularity and clarity}} within each sample, rather than merely the quantity of data, both of which are essential for effective instruction tuning.

From a visual feature perspective, the clarity of objects is often limited by the low-resolution of the data used in training. In 2D LMMs, a significant amount of research has been conducted on high-resolution images to enhance model performance by providing more detailed visual information~\cite{liu2024improved, liu2024llavanext, zhang2024llava-hd}. However, previous studies on 3D LMMs have predominantly used low-resolution point clouds~\cite{huang2023embodied,zhu20233d-vista}, which can lack sufficient detail, leading to ambiguous spatial representations. On the other hand, from a textual perspective, the data in existing 3D scene-text datasets often lack variety and detail, both critical for enhancing informational granularity and clarity, due to their construction using rule-based methods~\cite{azuma2022scanqa, chen2021scan2cap}. 
To mitigate the issue, we have focused on recent studies~\cite{zheng2023click, yan2024contrastive, sarkar2024mitigating} showing that negative samples help models distinguish more clearly between correct and incorrect responses.

In this work, we propose \textbf{CL3DOR}, \textbf{C}ontrastive \textbf{L}earning for \textbf{3D} large multimodal models via \textbf{O}dds ratio on high-\textbf{R}esolution point clouds, a novel approach that addresses quality issues in both visual and textual content data. As illustrated in Figure~\ref{fig:introduction}, we ensure greater specificity and clarity through two approaches. For visual features, we enhance the resolution by increasing the density of point clouds per object. For textual features, as shown in Figure~\ref{fig:datasets} (b), we use GPT-4o~\cite{openai2024gpt4o} to augment each response in the instruction tuning dataset with plausible hard negatives that the model could easily confuse. To utilize an augmented dataset, we reinterpret the objective function used in preference optimization, ORPO~\cite{hong2024reference}, from the perspective of contrastive learning by incorporating the odds ratio as an auxiliary component of the negative log-likelihood (NLL) loss, a process we refer to as spatial contrastive instruction tuning.

Applying spatial contrastive instruction tuning directly to CL3DOR, which uses an LLM as its backbone, poses challenges due to unaligned feature spaces with point clouds. To address the issue, we first conduct two alignment tuning stages for 3D-object and 3D-scene awareness before spatial contrastive instruction tuning. Consequently, as depicted in Figure~\ref{fig:datasets} (a), CL3DOR undergoes a three-stage training paradigm. Such a structured approach enables CL3DOR to fully exploit learnable information and significantly enhances cross-modal understanding by capturing fine-grained distinctions between positive and negative responses.

To validate the effectiveness of CL3DOR, we conduct extensive experiments across various benchmarks \cite{azuma2022scanqa, ma2022sqa3d, chen2021scan2cap}, assessing its spatial understanding and reasoning capabilities. The results demonstrate that CL3DOR achieves state-of-the-art performance on most metrics compared to task-adaptive models~\cite{hong20233d-llm, huang2023chat3dv2} and a general-purpose model~\cite{huang2023embodied} used as baselines. Additionally, on the 3D-object hallucination benchmark dataset~\cite{yang20243d-grand}, CL3DOR significantly outperforms the baseline models in both accuracy and F1 score. Moreover, through comprehensive experiments on resolution, negative dataset construction, and objective function, we experimentally demonstrate the importance and effectiveness of each component used in CL3DOR. The contributions of our work are threefold:
\begin{itemize}
    \item We propose CL3DOR which understands language instruction with point cloud data, using contrastive learning with the odds ratio on high-resolution point clouds. To the best of our knowledge, we are the first to redefine the objective function from preference optimization to adapt it for contrastive learning in 3D LMMs.
    \item We introduce methods to create high-quality training datasets for 3D LMMs from both visual and textual perspectives and release the constructed datasets.
    \item CL3DOR achieves state-of-the-art performance on 3D scene understanding and reasoning benchmarks, demonstrating its superiority in spatial understanding and reasoning. Also, we validate the effectiveness of each component of CL3DOR through extensive experiments.
\end{itemize}

\begin{figure*}[!ht]
\setlength{\abovecaptionskip}{0.2cm}
\setlength{\belowcaptionskip}{-0.2cm}
\centering
\includegraphics[width=17cm]{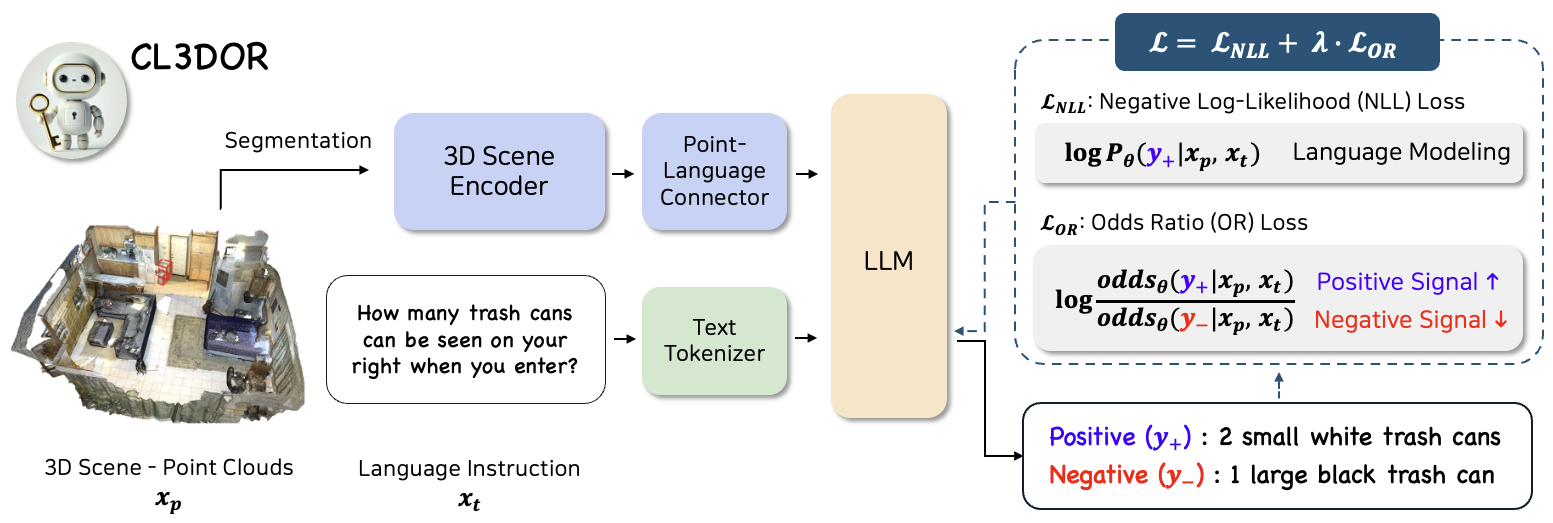}
\caption{Illustration of the proposed CL3DOR. The figure visualizes spatial contrastive instruction tuning. The objective function incorporates an odds ratio loss as an auxiliary term alongside the commonly used NLL loss for language modeling.}
\label{fig:overall_architecture}
\end{figure*}

\section{Related work}
\paragraph{3D Large Multimodal Model}
With the ongoing evolution of LLMs and 2D LMMs, research on 3D LMMs has significantly increased~\cite{wang2023chat3d,chen2024ll3da,qi2024gpt4point, zhou2023uni3d,fu2024scene, guo2023point}. 3D-LLM~\cite{hong20233d-llm} extracts 3D features from multi-view images and projects them into a pre-trained 2D LMM space, facilitating tasks such as 3D captioning, and question answering. However, it is constrained by limitations in semantic richness due to the inherent challenges of reconstructing 2D images into 3D features. In response, 3D-VisTA~\cite{zhu20233d-vista} uses self-attention layers for both single-modal modeling and multimodal fusion. Chat-3D v2~\cite{huang2023chat3dv2} employs unique object identifiers to enhance object referencing and scene understanding in 3D environments. LEO~\cite{huang2023embodied} presents a two-stage training model for an embodied multimodal generalist agent, achieving multi-task capabilities without additional fine-tuning. Despite these advancements, previous research has overlooked issues related to dataset quality, including the use of low-resolution point clouds and the inability to fully harness the rich information available in 3D scene data. These shortcomings restrict the models’ ability to achieve a comprehensive spatial representation.

\paragraph{Contrastive Learning with Language Generation}
Contrastive learning focuses on improving feature representation by decreasing the distance between positive pairs and increasing the distance between negative pairs. It has significantly influenced the field of natural language generation, with CoNT~\cite{an2022cont} aligning encoder and decoder representations for non-open-ended language generation, and BRIO~\cite{liu2022brio} using a contrastive loss to ensure the likelihood of sequences aligns with reference text similarity. Additionally, the exploration of contrastive learning related to sequence likelihood has progressed~\cite{jain2022contraclm}. Specifically, CLICK~\cite{zheng2023click} applies contrastive learning to sequence likelihood to avoid generating undesirable text, such as toxic language and unnatural repetition, whereas FDPO~\cite{gunjal2024detecting} and HALVA~\cite{sarkar2024mitigating} use it on sub-sentence level likelihood to mitigate hallucinations in 2D LMMs. To the best of our knowledge, CL3DOR is the first approach to apply contrastive learning to instruction tuning in the 3D LMMs.

\section{Methodology}
\subsection{Overview}
As depicted in Figure~\ref{fig:overall_architecture}, CL3DOR primarily consists of three components: the 3D scene encoder, the point-language connector, and the pre-trained Large Language Model (LLM). The 3D scene encoder, which includes a pre-trained point cloud encoder and a spatial transformer~\cite{chen2022language}, processes segmented 3D scene point clouds with 8,192 points per object for high-resolution. The extracted object-centric features are concatenated, transformed, and passed through the point-language connector, which uses MLP layers to generate object-specific tokens. These tokens, combined with tokenized language instructions, are input into the LLM to generate responses in an auto-regressive manner. Further details are provided in the Appendix.

\begin{table}[ht]
\centering
\caption{Dataset statistics for different training stages are provided. The `Type' column indicates whether the dataset is structured as (instruction, response) pairs or (instruction, positive response, negative response) triplets.}
\vspace{0.2cm}
\label{dataset_statistics}
\resizebox{\columnwidth}{!}{%
\begin{tabular}{llccc}
\toprule
\textbf{Stage} & \textbf{Task} & \textbf{Source} & \textbf{\# Data} & \textbf{Type}\\
\cmidrule[\heavyrulewidth]{1-5} 
Stage 1 & 3D-Object Captioning &  Cap3D & 660K & pair\\
\midrule
\multirow{2}{*}{Stage 2} & 3D-Object-in-the-Scene Captioning & ReferIt3D & 152K & pair \\
 & 3D Scene Captioning & Ours & 1.5K & pair \\
\midrule
\multirow{4}{*}{Stage 3} & 3D-Object-in-the-Scene Captioning & Scan2Cap & 37K & triplet \\
 & 3D Question Answering& ScanQA  & 25K  & triplet\\
 & 3D Question Answering&  SQA3D & 26K  & triplet\\
 & 3D-Object Existence  & Ours & 16K & triplet \\
\bottomrule
\end{tabular}%
}
\end{table}

\subsection{Training Paradigm in CL3DOR}
The training process of CL3DOR follows a three-stage paradigm, as outlined in~\cite{huang2023chat3dv2}. We include two alignment tuning stages, beginning with 3D-object awareness to learn about individual objects, and then expanding to 3D-scene awareness to grasp relationships among them. The final stage focuses on spatial contrastive instruction tuning to optimize performance in various downstream tasks. Detailed information about the training datasets used in each stage is provided in Table~\ref{dataset_statistics}. The likelihood of the output sequence \( y \) given the input \( x \) can be factorized autoregressively as follows:
\[
P_\theta(y \mid x) = \prod_{i=1}^{|y|} P_\theta(y^i \mid x, y^{<i}),
\]
where \( y^{<i} \) represents all tokens preceding \( y^i \). This factorization forms the basis for the Negative Log-Likelihood (NLL) loss, which is defined as:

\[
\mathcal{L}_{\text{NLL}} = -\frac{1}{{|y|}}\sum_{i=1}^{|y|} \log P_\theta(y^i \mid x_p, x_t, y^{<i}),
\]

\noindent where \( y \), \( x_p \), and \( x_t \) denote the label, 3D scene point clouds, and language instruction, respectively. The first and second stages only use NLL loss, while the third stage incorporates an additional term for contrastive learning.

\paragraph{Stage 1: 3D-Object Alignment Tuning}
The objective of the first stage is to enhance CL3DOR's understanding of 3D objects represented by point clouds. It is achieved by training the model on the Cap3D dataset~\cite{luo2024scalable}, which is derived from Objaverse~\cite{deitke2023objaverse}, and focuses on generating captions for 3D objects. In this stage, only the parameters of the spatial transformer and the point-language connector are updated.

\paragraph{Stage 2: Spatial Alignment Tuning}
In the second stage, CL3DOR extends its comprehension from individual objects to entire 3D scenes composed of multiple objects, thereby acquiring spatial awareness. The training involves the ReferIt3D dataset~\cite{achlioptas2020referit3d}, which emphasizes the description of referred objects. Also, a custom-built scene captioning dataset is used, generated by GPT-4o to produce captions based on multi-view screenshots of 3D scenes from various top-view angles. The trainable parameters remain consistent with those in the first stage.

\paragraph{Stage 3: Spatial Contrastive Instruction Tuning}
The final stage focuses on general downstream tasks such as 3D-object-in-the-scene captioning, question answering (QA), and grounded QA for object existence within 3D scenes. Unlike the previous stages, the final stage uses a triplet structure in the training dataset. To effectively leverage negative responses in triplet text data, we incorporate an auxiliary term \(\mathcal{L}_{\text{OR}}\), based on the odds calculated from the likelihood of generating the output sequence $y$ given an input sequence $x$, as shown in Eq.~\eqref{odds}. The process involves applying a logarithm to the ratio of the odds for positive (\(y_\text{+}\)) and negative (\(y_\text{-}\)) responses, followed by the sigmoid function \(\sigma\), as detailed in Eq.~\eqref{odds_ratio}. Notably, NLL loss is calculated only with the positive response $y_\text{+}$. The final objective function is defined as Eq.~\eqref{eq:combined_loss}, with $\lambda$ as a hyperparameter for the weighted term. During this phase, the trainable parameters include those of the spatial transformer, point-language connector, and LLM.

\begin{equation}
\text{odds}_{\theta}(y \mid x) = \frac{P_{\theta}(y \mid x)}{1 - P_{\theta}(y \mid x)}
\label{odds}
\end{equation}

\begin{equation}
\mathcal{L}_{\text{OR}} = -\log \sigma \left( \log \left( \frac{\text{odds}_{\theta}(y_\text{+} \mid x)}{\text{odds}_{\theta}(y_\text{-} \mid x)} \right) \right)
\label{odds_ratio}
\end{equation}

\begin{equation}
\mathcal{L} = \mathbb{E}_{(x, y_+, y_-)} \Big[\mathcal{L}_{\text{NLL}} + \lambda \cdot \mathcal{L}_{\text{OR}}\Big]
\label{eq:combined_loss}
\end{equation}

\begin{figure}[t]
\setlength{\abovecaptionskip}{0.2cm}
\setlength{\belowcaptionskip}{-0.2cm}
\centering
\includegraphics[width=7.3cm]{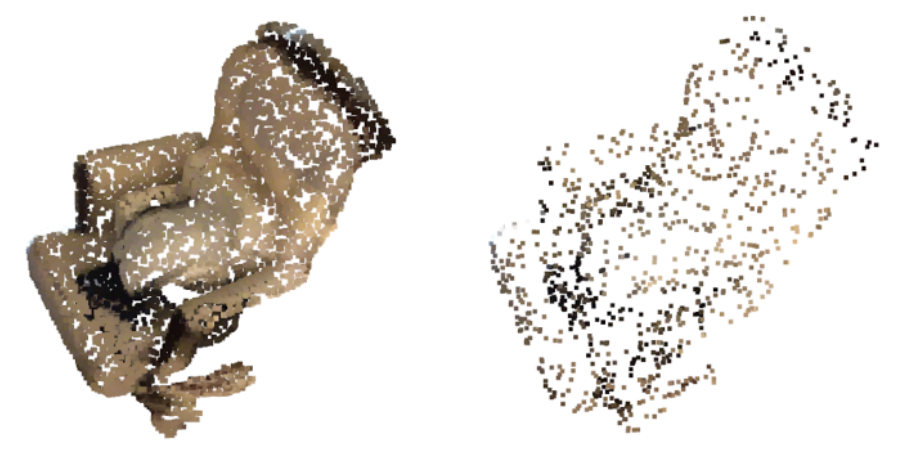}
\caption{Resolution in point clouds: high-resolution (left) with 8,192 point clouds per object, and low-resolution (right) with 1,024 point clouds per object.
}
\label{fig:resolution_ablation}
\end{figure}

\subsection{Dataset Refinement for CL3DOR}
\paragraph{High-Resolution Point Clouds}
Previous studies~\cite{zhu20233d-vista, huang2023embodied} typically sample 1,024 points per object for extracting features from 3D scenes. However, as illustrated in Figure~\ref{fig:resolution_ablation}, a resolution of 1,024 points results in significant information loss, making it challenging even for humans to discern object characteristics. Drawing on the importance of high-resolution visual content in 2D LMMs~\cite{liu2024improved, liu2024llavanext, zhang2024llava-hd}, we hypothesize that high-resolution point cloud sampling is crucial for accurate visual feature extraction.

To address the issue, we employ PointBERT~\cite{yu2022point}, a pre-trained point cloud encoder that processes 8,192 points per object to extract detailed 3D object features. The high sampling rate ensures the extraction of higher informational granularity of 3D object features, capturing even the finest details to enhance clarity and provide abundant information. We further conduct an ablation analysis comparing high-resolution and low-resolution sampling. The results, detailed in the Discussion section, highlight the importance of maintaining high point cloud density to reduce information loss and improve object representation fidelity.

\begin{table*}[!ht]
\centering
\caption{Quantitative comparison with state-of-the-art models for 3D scene understanding and reasoning tasks. In the `EM@1' column, values in parentheses represent refined exact-match scores. The `Sim' column denotes sentence similarity scores.
}
\vspace{0.2cm}
\label{main_table}
\resizebox{\textwidth}{!}{%
\begin{tabular}{lccccccccccc}
\toprule
& \multicolumn{5}{c}{ScanQA (val)} & SQA3D (test) & \multicolumn{5}{c}{Scan2Cap (val)} \\
\cmidrule(lr){2-6} \cmidrule(lr){7-7} \cmidrule(lr){8-12}
& CIDEr & BLEU-4 & METEOR & ROUGE-L & EM@1 & EM@1 & CIDEr & BLEU-4 & METEOR & ROUGE-L & Sim \\
\midrule
\textbf{Single-task models} \\
Scan2Cap & - & - & - & - & - & 41.0 & 35.2 & 22.4 & 21.4 & 43.5 & - \\
3DJCG & - & - & - & - & - & - & 47.7 & 31.5 & 24.3 & 51.8 & - \\
Vote2Cap-DETR & - & - & - & - & - & - & 61.8 & 34.5 & 26.2 & 54.4 & - \\
VoteNet+MCAN & 54.7 & 6.2 & 11.4 & 29.8 & 17.3 & - & - & - & - & - & - \\
ScanRefer+MCAN & 55.4 & 7.9 & 11.5 & 30.0 & 18.6 & - & - & - & - & - & - \\
ScanQA & 64.9 & 10.1 & 13.1 & 33.3 & 21.1 & 47.2 & - & - & - & - & - \\
ClipBERT & - & - & - & - & - & 43.3 & - & - & - & - & - \\
\midrule
\textbf{Task-adaptive models} \\
3D-VisTA & 69.6 & 10.4 & 13.9 & 35.7 & 22.4 & 48.5 & 66.9 & 34.0 & 27.1 & 54.3 & 53.8 \\
3D-LLM (Flamingo) & 59.2 & 7.2 & 12.2 & 32.3 & 20.4 & - & - & - & - & - & - \\
3D-LLM (FlanT5) & 69.4 & 12.0 & 14.5 & 35.7 & 20.5 & - & - & - & - & - & - \\
Chat-3D v2 & 77.1 & 7.3 & 16.1 & 40.1 & 21.1 & - & - & - & - & - & - \\
\midrule
\textbf{General-purpose models} \\
LEO & 101.4 & 13.2 & 20.0 & 49.2 & 24.5 (47.6) & 50.0 (52.4) & 72.4 & \textbf{38.2} & \textbf{27.9} & 58.1 & 55.3 \\
CL3DOR & \textbf{110.4} & \textbf{16.7} & \textbf{21.0} & \textbf{52.5} & \textbf{25.8} (\textbf{52.9}) & \textbf{51.6} (\textbf{54.4}) & \textbf{93.4} & 36.0 & 27.6 & \textbf{60.1} & \textbf{67.1} \\
\bottomrule
\end{tabular}%
}
\end{table*}

\paragraph{Hard Negative Response Generation}

For effective contrastive learning during the spatial contrastive instruction tuning of CL3DOR, it is essential to use triplet data comprising a question, a positive response, and a negative response. Inspired by prior studies~\cite{robinson2020contrastive, byun2022grit}, we incorporate hard negatives that are plausible yet incorrect, enabling the model to learn fine-grained features from a limited dataset efficiently. We devise pipelines that augment existing datasets with informative negatives generated by GPT-4o, excluding the 3D object existence task. Examples of the triplet data are shown in Figure~\ref{fig:datasets}, and detailed prompts for GPT-4o can be found in the Appendix.

For the 3D question-answering task, we generate hard negative responses for the ScanQA~\cite{azuma2022scanqa} and SQA3D~\cite{ma2022sqa3d} datasets. These datasets include questions about the presence, location, and attributes of specific objects in a 3D indoor scene. By leveraging a top-view image and a list of scene objects as inputs to GPT-4o, we craft contextually rich negatives. It ensures the granularity of the data, as hard negatives force the model to distinguish finer details and reduce ambiguity in responses.

Moreover, the 3D-object-in-the-scene captioning task generates concise descriptions of specified objects within a 3D scene, using the Scan2Cap~\cite{chen2021scan2cap} training dataset. To support contrastive learning, we augment the dataset with hard negative captions crafted by GPT-4o. Positive captions accurately describe the relative positions and attributes of the objects, while negative captions maintain the original structure but deliberately alter key elements such as location or attributes. This approach aims to sharpen the model's ability to distinguish subtle yet crucial differences.

Lastly, in the 3D-object existence task, we use ScanNet~\cite{dai2017scannet} data to create binary questions about object presence in a scene, with responses limited to \textit{yes} or \textit{no}. Unlike the tasks that use hard negatives to increase difficulty, the binary answer format limits the complexity of negative responses. To increase the difficulty of the questions, we generate \textit{no} responses by considering the frequency and co-occurrence of objects in the entire dataset.


\section{Experimental Setup}
\paragraph{Implementation details}

During the spatial contrastive instruction tuning, we set key hyperparameters as follows: one epoch, a learning rate of 6e-5 with a cosine annealing schedule, and a batch size of 64. The  \(\lambda\) for the OR loss is linearly ramped up from 0 to 3e-1 to ensure stable training, similar to the distillation weight \(\alpha\) in ALBEF~\cite{li2021albef}. We employ LLaMA3-8B-Instruct~\cite{dubey2024llama} as the LLM. All datasets used for training CL3DOR will be made publicly available. Further details can be found in the Appendix.

\paragraph{Benchmarks} We evaluate CL3DOR on various benchmarks, focusing on 3D scene understanding, reasoning, and object hallucination. We assess performance on four major benchmarks: 3D captioning on Scan2Cap~\cite{chen2021scan2cap}, 3D QA on ScanQA~\cite{azuma2022scanqa}, 3D embodied reasoning on SQA3D~\cite{ma2022sqa3d}, and 3D object hallucination on 3D-POPE~\cite{yang20243d-grand}. The 3D-POPE dataset, including \textit{Random}, \textit{Popular}, and \textit{Adversarial} settings in ScanNet v200~\cite{rozenberszki2022language}, identifies the presence or absence of objects within a scene.

\paragraph{Baselines} 
To evaluate CL3DOR's performance, we compare it against single-task models, task-adaptive models, and general-purpose models. Single-task models, trained solely to perform specific tasks, include Scan2Cap~\cite{chen2021scan2cap}, 3DJCG~\cite{cai20223djcg}, and Vote2Cap-DETR~\cite{Vote2Cap-DETR} for 3D captioning, as well as VoteNet (or ScanRefer) + MCAN~\cite{azuma2022scanqa}, CLIPBERT~\cite{lei2021clipbert}, and ScanQA~\cite{azuma2022scanqa} for 3D QA. Task-adaptive models, which are fine-tuned for specific tasks after the pre-training stage, include 3D-LLM (Flamingo, FlanT5)~\cite{hong20233d-llm}, 3D-VisTA~\cite{zhu20233d-vista}, and Chat-3D v2~\cite{huang2023chat3dv2}. The only baseline for general-purpose models, capable of performing various tasks without task-specific fine-tuning, is LEO~\cite{huang2023embodied}.

\paragraph{Evaluation Metrics} For the evaluation metrics, we employ BLEU~\cite{papineni2002bleu}, ROUGE~\cite{lin2004rouge}, METEOR~\cite{banerjee2005meteor}, CIDEr~\cite{vedantam2015cider}, and refined exact-match accuracy, with the latter referenced from~\cite{huang2023embodied} to ensure fair comparisons. Also, we use sentence similarity~\cite{reimers2019sentence} for open-ended generation evaluation.

\section{Experimental Results}
\subsection{Main Results}

As shown in Table~\ref{main_table}, CL3DOR achieves state-of-the-art performance across most metrics in all 3D scene understanding and reasoning datasets. Despite single-task models and task-adaptive models being specifically fine-tuned for particular tasks, CL3DOR, a general-purpose model, delivers the highest performance without any task-specific fine-tuning. While the previous state-of-the-art general-purpose model, LEO, also performs well, CL3DOR surpasses it significantly. Specifically, in ScanQA, CL3DOR outperforms LEO across all metrics, with notable improvements of 9 points in CIDEr and 5.3\%p in EM@1-refined. Similarly, in the Scan2Cap dataset, CL3DOR exceeds LEO by 21 points in CIDEr and by 11.8\%p in sentence similarity scores. While CL3DOR shows lower BLEU-4 and METEOR scores on Scan2Cap, these metrics heavily emphasize lexical similarity and thus fail to fully capture the model's performance. As noted in Chat-3D v2, the higher CIDEr and sentence similarity scores suggest that our model produces more diverse and semantically rich outputs, potentially capturing the essence of the scene more effectively.

\begin{table}[t]
\centering
\caption{3D-POPE benchmark results for evaluating 3D object hallucination on ScanNet200 Val. The \textit{Yes} column shows the percentage of \textit{yes} responses among \textit{yes} and \textit{no} responses. The `\#Size' column indicates the number of 3D scene-text samples in the training datasets. Rows shaded in gray represent models trained on a large-scale 6.8M dataset and are not directly compared with CL3DOR.}
\vspace{0.2cm}
\label{pope_table}
\resizebox{\columnwidth}{!}{%
\begin{tabular}{llcccc}
\toprule
\textbf{Dataset} & \textbf{Model} & \textbf{\#Size} & \textbf{F1 Score} & \textbf{Accuracy} & \textbf{Yes (\%)} \\
\midrule
\multirow{6}{*}{\textbf{\textit{Random}}} & Random Baseline & - & 50.0 & 50.0 & 50.00 \\
& 3D-LLM & 300K & 66.7 & 50.1 & 99.8 \\
& 3D-VisTA & 278K & 51.8 & 49.7 & 54.0 \\
& LEO & 519K & 62.3 & 52.9 & 74.7 \\
& CL3DOR & 257.5K & \textbf{75.9} & \textbf{76.9} & 54.1 \\
& \cellcolor{gray!30}3D-GRAND & \cellcolor{gray!30} 6.8M & \cellcolor{gray!30} 88.6 & \cellcolor{gray!30} 89.1 & \cellcolor{gray!30} 45.1 \\
\midrule
\multirow{6}{*}{\textbf{\textit{Popular}}} & Random Baseline & - & 50.0 & 50.0 & 50.0 \\
& 3D-LLM & 300K & \textbf{66.6} & 49.9 & 99.9 \\
& 3D-VisTA & 278K & 49.5 & 49.5 & 52.3 \\
& LEO & 519K & 59.6 & 47.3 & 80.4 \\
& CL3DOR & 257.5K & 65.0 & \textbf{69.6} & 65.0 \\
& \cellcolor{gray!30}3D-GRAND & \cellcolor{gray!30} 6.8M & \cellcolor{gray!30} 78.3 & \cellcolor{gray!30} 76.6 & \cellcolor{gray!30} 57.7 \\
\midrule
\multirow{6}{*}{\textbf{\textit{Adversarial}}} & Random Baseline & - & 50.0 & 50.0 & 50.0 \\
& 3D-LLM & 300K & 66.6 & 49.9 & 99.9 \\
& 3D-VisTA & 278K & 51.2 & 51.1 & 53.0 \\
& LEO & 519K & 59.8 & 47.5 & 80.5 \\
& CL3DOR & 257.5K & \textbf{68.1} & \textbf{62.2} & 68.5 \\
& \cellcolor{gray!30}3D-GRAND & \cellcolor{gray!30} 6.8M & \cellcolor{gray!30} 76.4 & \cellcolor{gray!30} 74.0 & \cellcolor{gray!30} 60.3 \\
\bottomrule
\end{tabular}%
}
\end{table}

Furthermore, we evaluate CL3DOR on the 3D object hallucination task. Table~\ref{pope_table} shows CL3DOR's performance on the 3D-POPE benchmark under \textit{Random}, \textit{Popular}, and \textit{Adversarial} settings. We do not directly compare our method with 3D-GRAND~\cite{yang20243d-grand}, as it is trained on a substantially larger dataset containing 6.8M 3D scene-text samples; even for the existence task alone, it uses 532K samples, further highlighting the disparity in dataset sizes. Therefore, rather than comparing absolute performance metrics, we focus on comparisons with models trained on a relatively similar number of samples to ensure a fair assessment of CL3DOR's object hallucination capabilities. The baselines 3D-LLM, 3D-VisTA, and LEO exhibit similar accuracy levels to the random baseline. Notably, LEO and 3D-LLM show high \textit{Yes} rates, indicating a severe bias towards the existence of objects. In contrast, CL3DOR maintains a more stable \textit{Yes} rate while achieving higher F1 scores and accuracy compared to the baselines. These findings indicate that CL3DOR not only excels in general 3D scene understanding and reasoning tasks but also demonstrates robust performance in 3D object hallucination tasks.

\subsection{Impact of High-Quality Visual and Textual Data}

\paragraph{Superior Results with Higher Object Resolution.}
Table~\ref{ablation_table} presents the performance of CL3DOR trained under both high-resolution and low-resolution settings, sampling 8,192 and 1,024 point clouds per object, respectively, on the ScanQA, Scan2Cap, and 3D-POPE benchmarks. In all cases, the high-resolution settings result in superior performance across all datasets and metrics compared to the low-resolution settings. Particularly in the Scan2Cap dataset, the most significant improvement is observed in the CIDEr metric, indicating that higher resolution substantially enhances the model’s capability to generate precise captions for specific objects. Furthermore, the increase in computational cost due to the larger size of the point cloud is not significant because the point cloud encoder within the 3D scene encoder accounts for only 0.27\% of CL3DOR's total parameters. Most of the computational load is handled by the LLM, so the increase in computation cost does not significantly impact the system’s efficiency.


\begin{table}[t]
\centering
\caption{Ablation studies investigating the performance of high-resolution data and hard negative responses. `w/ Low Res.' indicates training with low-resolution data, and `w/ Easy Neg.' indicates training with easy negative responses.}
\vspace{0.2cm}
\label{ablation_table}
\resizebox{\columnwidth}{!}{%
\begin{tabular}{llcccc}
\toprule
\textbf{Dataset} & \textbf{Metric} & \textbf{CL3DOR} & \textbf{w/ Low Res.} & \textbf{w/ Easy Neg.} \\
\midrule
\multirow{6}{*}{\textbf{ScanQA}} & CIDEr & \textbf{110.4} & 97.6 & 106.0\\
& BLEU-4 & \textbf{16.7} & 14.5 & 14.5\\
& METEOR & \textbf{21.0} & 18.9 & 20.5\\
& ROUGE-L & \textbf{52.5} & 47.5 & 51.7\\
& EM@1-refined & \textbf{52.9} & 45.8 & 51.7\\
\midrule
\multirow{6}{*}{\textbf{Scan2Cap}} & CIDEr & \textbf{93.4} & 71.7 & 92.6\\
& BLEU-4 & \textbf{36.0} & 33.5 & 35.8\\
& METEOR & \textbf{27.6} & 26.1 & 27.5\\
& ROUGE-L & 60.1 & 58.3 & \textbf{60.2}\\
& Sim & \textbf{67.1} & 61.1 & {67.0}\\
\midrule
\multirow{3}{*}{\makecell{\textbf{3D-POPE} \\ (\textit{Random})}} 
& F1 Score & \textbf{75.9} & 69.6 & 71.4\\
& Accuracy & \textbf{76.9} & 69.9 & 72.4\\
& Yes (\%) & 54.1 & 49.0 & 53.5\\
\midrule
\multirow{3}{*}{\makecell{\textbf{3D-POPE} \\ (\textit{Popular})}} 
& F1 Score & \textbf{65.0} & 62.0 & {63.4}\\
& Accuracy & \textbf{69.6} & 57.9 & 67.2\\
& Yes (\%) & 65.0 & 61.0 & 61.6\\
\midrule
\multirow{3}{*}{\makecell{\textbf{3D-POPE} \\ (\textit{Adversarial})}} 
& F1 Score & \textbf{68.1} & 60.1 & 65.8\\
& Accuracy & 62.2 & 57.2 & \textbf{65.5}\\
& Yes (\%) & 68.5 & 62.4 & 63.6\\
\bottomrule
\end{tabular}%
}
\end{table}

\begin{figure*}[!ht]
\setlength{\abovecaptionskip}{0.2cm}
\setlength{\belowcaptionskip}{-0.2cm}
\centering
\includegraphics[width=16.8cm]{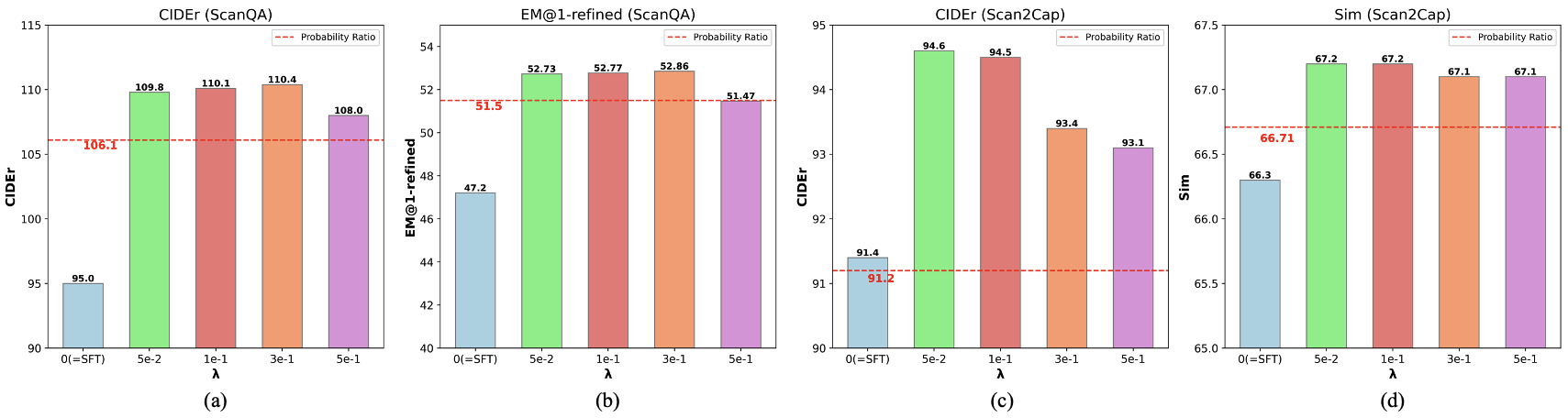}
\caption{Impact of $\lambda$ on Performance. Plots (a) and (b) illustrate the CIDEr and EM@1-refined scores for ScanQA, while plots (c) and (d) depict the CIDEr and sentence similarity (Sim) scores for Scan2Cap across varying $\lambda$ values. These results emphasize the crucial role of the OR term, $\lambda$, in enhancing performance, particularly when contrasted with the baseline of supervised fine-tuning (SFT) using only positive responses ($\lambda = 0$). The dashed red line represents the performance of CL3DOR when utilizing the probability ratio of sequence likelihood as an auxiliary term ($\lambda =$ 3e-1) instead of the odds ratio.}
\label{fig:lambda_ablation}
\end{figure*}

\begin{figure}[htbp]
\setlength{\abovecaptionskip}{0.2cm}
\setlength{\belowcaptionskip}{-0.2cm}
\centering
\includegraphics[width=\columnwidth]{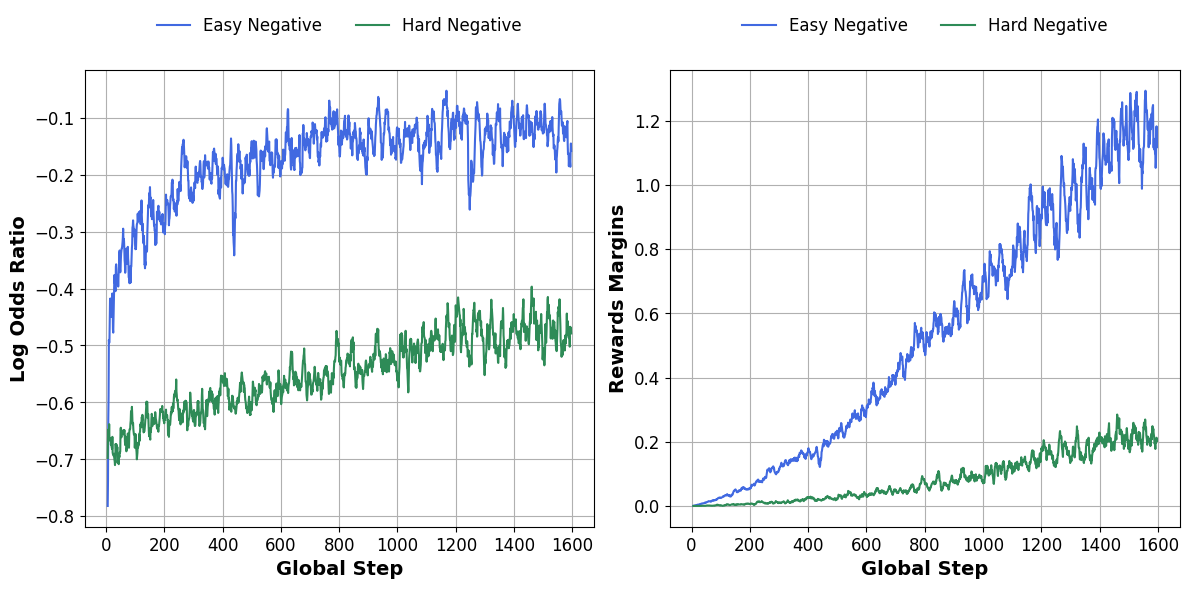}
\caption{Visualization of the log odds ratio (left) and the reward margin (right) for both hard negative and easy negative dataset settings. The blue line shows results using hard negatives, while the green line shows those using easy negatives.}
\label{fig:hard_easy_plot}
\end{figure}

\paragraph{The Crucial Role of Hard Negatives in Spatial Contrastive Instruction Tuning.}
To verify the effectiveness of the hard negative dataset constructed during spatial contrastive instruction tuning, we create an additional easy negative dataset and conduct an ablation study based on the difficulty of the negative datasets. The easy negative settings expand the original instruction-response pairs into triplets by incorporating responses randomly selected from different instructions within the dataset. As shown in Table~\ref{ablation_table}, the experimental results consistently demonstrate that the hard negative settings yield higher performance in most benchmarks compared to the easy negative settings.

Moreover, we analyze the log odds ratio represented as \(\log \left( \frac{\text{odds}_{\theta}(y_{+} \mid x)}{\text{odds}_{\theta}(y_{-} \mid x)} \right)\), and the reward margin, defined as the difference between the log probabilities of positive and negative responses, for both easy and hard negatives during contrastive learning. The analyses, depicted in Figure~\ref{fig:hard_easy_plot}, reveal that CL3DOR trained with hard negatives (green line) consistently exhibits lower log odds ratios and reward margins compared to those trained with easy negatives (blue line) throughout most training steps. These metrics suggest that hard negatives pose a greater challenge to the model, thereby driving the development of a more nuanced and robust understanding in context. Consequently, the hard negative setting fosters deeper learning and better equips the model to fully harness the information in the data for improving spatial understanding, as evidenced by the experimental results.

\subsection{Impact of Odds Ratio in Objective Function}
\paragraph{Stability and Performance: Why Odds Ratio Excels?}
Following previous work~\cite{hong2024reference}, we compare two objective functions for spatial contrastive instruction tuning: NLL loss with either the probability ratio of sequence likelihood,  $-[\log P_{\theta}(y_\text{+} | x) - \log P_{\theta}(y_\text{-} | x)]$, or the odds ratio (OR) loss. Our results, shown in Figure~\ref{fig:lambda_ablation}, demonstrate that while incorporating negative signals using the probability ratio loss (red dotted line) yields better performance than NLL loss (blue bar) alone, the OR loss provides even better overall performance. The rationale for these findings can be traced back to the discussion in~\cite{hong2024reference}, where it is noted that the odds ratio is more stable and avoids extreme discrimination against negative responses, maintaining a balance that prevents overly suppressing their logits. Such a balance is crucial in combining NLL and contrastive learning, as the odds ratio prevents degeneration by offering moderate contrast between positive and negative examples. Thus, the odds ratio is a better choice for NLL than the probability ratio, ensuring mild discrimination of negative responses while prioritizing positive responses.

\paragraph{Exploring $\lambda$ through Ablation Studies.}

To evaluate the impact of the odds ratio term in the objective function, we conduct an ablation study by varying the hyperparameter 
$\lambda$ across different values: 0, 5e-2, 1e-1, 3e-1, and 5e-1. When $\lambda$ is set to 0, the training setup is equivalent to supervised fine-tuning (SFT) with only positive responses, referred to as the original label. As shown in Figure~\ref{fig:lambda_ablation}, performance significantly drops at $\lambda = 0$ compared to other settings, and notably, the SFT case at $\lambda=0$ even performs worse than the CL3DOR trained with easy negative settings, as demonstrated in Table~\ref{ablation_table}. Specifically, setting $\lambda$ to 3e-1 results in a significant improvement in CIDEr scores, with an increase of 15.4 points for ScanQA and 2.0 points for Scan2Cap, compared to setting $\lambda$ to 0. The results underscore the importance of incorporating negative responses to enhance the model's spatial understanding. Moreover, the consistent performance across different $\lambda$ values indicates the robustness of our method with respect to the hyperparameter.

\section{Conclusion}
In this work, we propose CL3DOR, a 3D Large Multimodal Model designed to effectively harness high informational granularity and clarity in both visual and textual content. We enhance visual features by increasing the density of point clouds per object, thereby reducing information loss and achieving high-fidelity visual representations. For textual features, we augment the instruction tuning dataset with plausible hard negative responses to provide more meaningful information. To fully leverage the expanded dataset with hard negatives, we apply contrastive learning by incorporating an odds ratio term into the NLL loss, reinterpreting the objective function used in preference optimization. Consequently, CL3DOR achieves state-of-the-art performance on 3D scene understanding and reasoning benchmarks, surpassing existing baselines.

\paragraph{Limitation} Despite these promising results, our evaluation is constrained by limited resources, allowing only few epochs of training and precluding the use of million-scale 3D scene-text samples. These limitations prevent us from fully exploring potential performance improvements with extended training. We will address these constraints in future research.

\bibliography{main}

\begin{thebibliography}{61}
\providecommand{\natexlab}[1]{#1}
\providecommand{\url}[1]{\texttt{#1}}
\expandafter\ifx\csname urlstyle\endcsname\relax
  \providecommand{\doi}[1]{doi: #1}\else
  \providecommand{\doi}{doi: \begingroup \urlstyle{rm}\Url}\fi

\bibitem[Achlioptas et~al.(2020)Achlioptas, Abdelreheem, Xia, Elhoseiny, and Guibas]{achlioptas2020referit3d}
Achlioptas, P., Abdelreheem, A., Xia, F., Elhoseiny, M., and Guibas, L.
\newblock Referit3d: Neural listeners for fine-grained 3d object identification in real-world scenes.
\newblock In \emph{Computer Vision--ECCV 2020: 16th European Conference, Glasgow, UK, August 23--28, 2020, Proceedings, Part I 16}, pp.\  422--440. Springer, 2020.

\bibitem[An et~al.(2022)An, Feng, Lv, Kong, Qiu, and Huang]{an2022cont}
An, C., Feng, J., Lv, K., Kong, L., Qiu, X., and Huang, X.
\newblock Cont: Contrastive neural text generation.
\newblock \emph{Advances in Neural Information Processing Systems}, 35:\penalty0 2197--2210, 2022.

\bibitem[Azuma et~al.(2022)Azuma, Miyanishi, Kurita, and Kawanabe]{azuma2022scanqa}
Azuma, D., Miyanishi, T., Kurita, S., and Kawanabe, M.
\newblock Scanqa: 3d question answering for spatial scene understanding.
\newblock In \emph{proceedings of the IEEE/CVF conference on computer vision and pattern recognition}, pp.\  19129--19139, 2022.

\bibitem[Bai et~al.(2023)Bai, Bai, Yang, Wang, Tan, Wang, Lin, Zhou, and Zhou]{bai2023qwen}
Bai, J., Bai, S., Yang, S., Wang, S., Tan, S., Wang, P., Lin, J., Zhou, C., and Zhou, J.
\newblock Qwen-vl: A frontier large vision-language model with versatile abilities.
\newblock \emph{arXiv preprint arXiv:2308.12966}, 2023.

\bibitem[Banerjee \& Lavie(2005)Banerjee and Lavie]{banerjee2005meteor}
Banerjee, S. and Lavie, A.
\newblock Meteor: An automatic metric for mt evaluation with improved correlation with human judgments.
\newblock In \emph{Proceedings of the acl workshop on intrinsic and extrinsic evaluation measures for machine translation and/or summarization}, pp.\  65--72, 2005.

\bibitem[Byun et~al.(2022)Byun, Hwang, Fu, and Moon]{byun2022grit}
Byun, J., Hwang, T., Fu, J., and Moon, T.
\newblock Grit-vlp: Grouped mini-batch sampling for efficient vision and language pre-training.
\newblock In \emph{European Conference on Computer Vision}, pp.\  395--412. Springer, 2022.

\bibitem[Cai et~al.(2022)Cai, Zhao, Zhang, Sheng, and Xu]{cai20223djcg}
Cai, D., Zhao, L., Zhang, J., Sheng, L., and Xu, D.
\newblock 3djcg: A unified framework for joint dense captioning and visual grounding on 3d point clouds.
\newblock In \emph{Proceedings of the IEEE/CVF Conference on Computer Vision and Pattern Recognition}, pp.\  16464--16473, 2022.

\bibitem[Cha et~al.(2024)Cha, Kang, Mun, and Roh]{cha2024honeybee}
Cha, J., Kang, W., Mun, J., and Roh, B.
\newblock Honeybee: Locality-enhanced projector for multimodal llm.
\newblock In \emph{Proceedings of the IEEE/CVF Conference on Computer Vision and Pattern Recognition}, pp.\  13817--13827, 2024.

\bibitem[Chang et~al.(2015)Chang, Funkhouser, Guibas, Hanrahan, Huang, Li, Savarese, Savva, Song, Su, et~al.]{chang2015shapenet}
Chang, A.~X., Funkhouser, T., Guibas, L., Hanrahan, P., Huang, Q., Li, Z., Savarese, S., Savva, M., Song, S., Su, H., et~al.
\newblock Shapenet: An information-rich 3d model repository.
\newblock \emph{arXiv preprint arXiv:1512.03012}, 2015.

\bibitem[Chen et~al.(2024{\natexlab{a}})Chen, Wu, Chen, Liu, He, Xiong, Liu, Guo, and Huang]{chen2024your}
Chen, R., Wu, Y., Chen, L., Liu, G., He, Q., Xiong, T., Liu, C., Guo, J., and Huang, H.
\newblock Your vision-language model itself is a strong filter: Towards high-quality instruction tuning with data selection.
\newblock \emph{arXiv preprint arXiv:2402.12501}, 2024{\natexlab{a}}.

\bibitem[Chen et~al.(2022)Chen, Guhur, Tapaswi, Schmid, and Laptev]{chen2022language}
Chen, S., Guhur, P.-L., Tapaswi, M., Schmid, C., and Laptev, I.
\newblock Language conditioned spatial relation reasoning for 3d object grounding.
\newblock \emph{Advances in neural information processing systems}, 35:\penalty0 20522--20535, 2022.

\bibitem[Chen et~al.(2023)Chen, Zhu, Chen, Lei, Yu, and Chen]{Vote2Cap-DETR}
Chen, S., Zhu, H., Chen, X., Lei, Y., Yu, G., and Chen, T.
\newblock End-to-end 3d dense captioning with vote2cap-detr.
\newblock In \emph{Proceedings of the IEEE/CVF Conference on Computer Vision and Pattern Recognition}, pp.\  11124--11133, 2023.

\bibitem[Chen et~al.(2024{\natexlab{b}})Chen, Chen, Zhang, Li, Yu, Fei, Zhu, Fan, and Chen]{chen2024ll3da}
Chen, S., Chen, X., Zhang, C., Li, M., Yu, G., Fei, H., Zhu, H., Fan, J., and Chen, T.
\newblock Ll3da: Visual interactive instruction tuning for omni-3d understanding reasoning and planning.
\newblock In \emph{Proceedings of the IEEE/CVF Conference on Computer Vision and Pattern Recognition}, pp.\  26428--26438, 2024{\natexlab{b}}.

\bibitem[Chen et~al.(2021)Chen, Gholami, Nie{\ss}ner, and Chang]{chen2021scan2cap}
Chen, Z., Gholami, A., Nie{\ss}ner, M., and Chang, A.~X.
\newblock Scan2cap: Context-aware dense captioning in rgb-d scans.
\newblock In \emph{Proceedings of the IEEE/CVF conference on computer vision and pattern recognition}, pp.\  3193--3203, 2021.

\bibitem[Dai et~al.(2017)Dai, Chang, Savva, Halber, Funkhouser, and Nie{\ss}ner]{dai2017scannet}
Dai, A., Chang, A.~X., Savva, M., Halber, M., Funkhouser, T., and Nie{\ss}ner, M.
\newblock Scannet: Richly-annotated 3d reconstructions of indoor scenes.
\newblock In \emph{Proceedings of the IEEE conference on computer vision and pattern recognition}, pp.\  5828--5839, 2017.

\bibitem[Deitke et~al.(2023)Deitke, Schwenk, Salvador, Weihs, Michel, VanderBilt, Schmidt, Ehsani, Kembhavi, and Farhadi]{deitke2023objaverse}
Deitke, M., Schwenk, D., Salvador, J., Weihs, L., Michel, O., VanderBilt, E., Schmidt, L., Ehsani, K., Kembhavi, A., and Farhadi, A.
\newblock Objaverse: A universe of annotated 3d objects.
\newblock In \emph{Proceedings of the IEEE/CVF Conference on Computer Vision and Pattern Recognition}, pp.\  13142--13153, 2023.

\bibitem[Dubey et~al.(2024)Dubey, Jauhri, Pandey, Kadian, Al-Dahle, Letman, Mathur, Schelten, Yang, Fan, et~al.]{dubey2024llama}
Dubey, A., Jauhri, A., Pandey, A., Kadian, A., Al-Dahle, A., Letman, A., Mathur, A., Schelten, A., Yang, A., Fan, A., et~al.
\newblock The llama 3 herd of models.
\newblock \emph{arXiv preprint arXiv:2407.21783}, 2024.

\bibitem[Fu et~al.(2024)Fu, Liu, Chen, Nie, and Xiong]{fu2024scene}
Fu, R., Liu, J., Chen, X., Nie, Y., and Xiong, W.
\newblock Scene-llm: Extending language model for 3d visual understanding and reasoning.
\newblock \emph{arXiv preprint arXiv:2403.11401}, 2024.

\bibitem[Gunjal et~al.(2024)Gunjal, Yin, and Bas]{gunjal2024detecting}
Gunjal, A., Yin, J., and Bas, E.
\newblock Detecting and preventing hallucinations in large vision language models.
\newblock In \emph{Proceedings of the AAAI Conference on Artificial Intelligence}, volume~38, pp.\  18135--18143, 2024.

\bibitem[Guo et~al.(2023)Guo, Zhang, Zhu, Tang, Ma, Han, Chen, Gao, Li, Li, et~al.]{guo2023point}
Guo, Z., Zhang, R., Zhu, X., Tang, Y., Ma, X., Han, J., Chen, K., Gao, P., Li, X., Li, H., et~al.
\newblock Point-bind \& point-llm: Aligning point cloud with multi-modality for 3d understanding, generation, and instruction following.
\newblock \emph{arXiv preprint arXiv:2309.00615}, 2023.

\bibitem[Hendrycks \& Gimpel(2016)Hendrycks and Gimpel]{hendrycks2016gaussian}
Hendrycks, D. and Gimpel, K.
\newblock Gaussian error linear units (gelus).
\newblock \emph{arXiv preprint arXiv:1606.08415}, 2016.

\bibitem[Hong et~al.(2024)Hong, Lee, and Thorne]{hong2024reference}
Hong, J., Lee, N., and Thorne, J.
\newblock Reference-free monolithic preference optimization with odds ratio.
\newblock \emph{arXiv preprint arXiv:2403.07691}, 2024.

\bibitem[Hong et~al.(2023)Hong, Zhen, Chen, Zheng, Du, Chen, and Gan]{hong20233d-llm}
Hong, Y., Zhen, H., Chen, P., Zheng, S., Du, Y., Chen, Z., and Gan, C.
\newblock 3d-llm: Injecting the 3d world into large language models.
\newblock \emph{Advances in Neural Information Processing Systems}, 36:\penalty0 20482--20494, 2023.

\bibitem[Huang et~al.(2023)Huang, Wang, Huang, Liu, Cheng, Zhao, Jin, and Zhao]{huang2023chat3dv2}
Huang, H., Wang, Z., Huang, R., Liu, L., Cheng, X., Zhao, Y., Jin, T., and Zhao, Z.
\newblock Chat-3d v2: Bridging 3d scene and large language models with object identifiers.
\newblock \emph{arXiv preprint arXiv:2312.08168}, 2023.

\bibitem[Huang et~al.(2024)Huang, Yong, Ma, Linghu, Li, Wang, Li, Zhu, Jia, and Huang]{huang2023embodied}
Huang, J., Yong, S., Ma, X., Linghu, X., Li, P., Wang, Y., Li, Q., Zhu, S.-C., Jia, B., and Huang, S.
\newblock An embodied generalist agent in 3d world.
\newblock In \emph{Proceedings of the International Conference on Machine Learning (ICML)}, 2024.

\bibitem[Jain et~al.(2022)Jain, Zhang, Ahmad, Wang, Nan, Li, Tan, Nallapati, Ray, Bhatia, et~al.]{jain2022contraclm}
Jain, N., Zhang, D., Ahmad, W.~U., Wang, Z., Nan, F., Li, X., Tan, M., Nallapati, R., Ray, B., Bhatia, P., et~al.
\newblock Contraclm: Contrastive learning for causal language model.
\newblock \emph{arXiv preprint arXiv:2210.01185}, 2022.

\bibitem[Jian et~al.(2024)Jian, Gao, and Vosoughi]{jian2024bootstrapping}
Jian, Y., Gao, C., and Vosoughi, S.
\newblock Bootstrapping vision-language learning with decoupled language pre-training.
\newblock \emph{Advances in Neural Information Processing Systems}, 36, 2024.

\bibitem[Jiang et~al.(2023)Jiang, Sablayrolles, Mensch, Bamford, Chaplot, Casas, Bressand, Lengyel, Lample, Saulnier, et~al.]{jiang2023mistral}
Jiang, A.~Q., Sablayrolles, A., Mensch, A., Bamford, C., Chaplot, D.~S., Casas, D. d.~l., Bressand, F., Lengyel, G., Lample, G., Saulnier, L., et~al.
\newblock Mistral 7b.
\newblock \emph{arXiv preprint arXiv:2310.06825}, 2023.

\bibitem[Lei et~al.(2021)Lei, Li, Zhou, Gan, Berg, Bansal, and Liu]{lei2021clipbert}
Lei, J., Li, L., Zhou, L., Gan, Z., Berg, T.~L., Bansal, M., and Liu, J.
\newblock Less is more: Clipbert for video-and-language learning via sparse sampling.
\newblock In \emph{Proceedings of the IEEE/CVF conference on computer vision and pattern recognition}, pp.\  7331--7341, 2021.

\bibitem[Li et~al.(2021)Li, Selvaraju, Gotmare, Joty, Xiong, and Hoi]{li2021albef}
Li, J., Selvaraju, R., Gotmare, A., Joty, S., Xiong, C., and Hoi, S. C.~H.
\newblock Align before fuse: Vision and language representation learning with momentum distillation.
\newblock \emph{Advances in neural information processing systems}, 34:\penalty0 9694--9705, 2021.

\bibitem[Lin(2004)]{lin2004rouge}
Lin, C.-Y.
\newblock Rouge: A package for automatic evaluation of summaries.
\newblock In \emph{Text summarization branches out}, pp.\  74--81, 2004.

\bibitem[Liu et~al.(2024{\natexlab{a}})Liu, Li, Li, and Lee]{liu2024improved}
Liu, H., Li, C., Li, Y., and Lee, Y.~J.
\newblock Improved baselines with visual instruction tuning.
\newblock In \emph{Proceedings of the IEEE/CVF Conference on Computer Vision and Pattern Recognition}, pp.\  26296--26306, 2024{\natexlab{a}}.

\bibitem[Liu et~al.(2024{\natexlab{b}})Liu, Li, Li, Li, Zhang, Shen, and Lee]{liu2024llavanext}
Liu, H., Li, C., Li, Y., Li, B., Zhang, Y., Shen, S., and Lee, Y.~J.
\newblock Llava-next: Improved reasoning, ocr, and world knowledge, January 2024{\natexlab{b}}.
\newblock URL \url{https://llava-vl.github.io/blog/2024-01-30-llava-next/}.

\bibitem[Liu et~al.(2024{\natexlab{c}})Liu, Li, Wu, and Lee]{liu2024visual}
Liu, H., Li, C., Wu, Q., and Lee, Y.~J.
\newblock Visual instruction tuning.
\newblock \emph{Advances in neural information processing systems}, 36, 2024{\natexlab{c}}.

\bibitem[Liu et~al.(2022)Liu, Liu, Radev, and Neubig]{liu2022brio}
Liu, Y., Liu, P., Radev, D., and Neubig, G.
\newblock Brio: Bringing order to abstractive summarization.
\newblock \emph{arXiv preprint arXiv:2203.16804}, 2022.

\bibitem[Luo et~al.(2024)Luo, Rockwell, Lee, and Johnson]{luo2024scalable}
Luo, T., Rockwell, C., Lee, H., and Johnson, J.
\newblock Scalable 3d captioning with pretrained models.
\newblock \emph{Advances in Neural Information Processing Systems}, 36, 2024.

\bibitem[Ma et~al.(2022)Ma, Yong, Zheng, Li, Liang, Zhu, and Huang]{ma2022sqa3d}
Ma, X., Yong, S., Zheng, Z., Li, Q., Liang, Y., Zhu, S.-C., and Huang, S.
\newblock Sqa3d: Situated question answering in 3d scenes.
\newblock \emph{arXiv preprint arXiv:2210.07474}, 2022.

\bibitem[OpenAI(2024)]{openai2024gpt4o}
OpenAI.
\newblock Gpt-4o.
\newblock \url{https://openai.com/index/hello-gpt-4o/}, 2024.

\bibitem[Papineni et~al.(2002)Papineni, Roukos, Ward, and Zhu]{papineni2002bleu}
Papineni, K., Roukos, S., Ward, T., and Zhu, W.-J.
\newblock Bleu: a method for automatic evaluation of machine translation.
\newblock In \emph{Proceedings of the 40th annual meeting of the Association for Computational Linguistics}, pp.\  311--318, 2002.

\bibitem[Patil et~al.(2023)Patil, Vasu, and Srinadh]{patil2023advances}
Patil, S., Vasu, V., and Srinadh, K.
\newblock Advances and perspectives in collaborative robotics: a review of key technologies and emerging trends.
\newblock \emph{Discover Mechanical Engineering}, 2\penalty0 (1):\penalty0 13, 2023.

\bibitem[Qi et~al.(2024)Qi, Fang, Sun, Wu, Wu, Wang, Lin, and Zhao]{qi2024gpt4point}
Qi, Z., Fang, Y., Sun, Z., Wu, X., Wu, T., Wang, J., Lin, D., and Zhao, H.
\newblock Gpt4point: A unified framework for point-language understanding and generation.
\newblock In \emph{Proceedings of the IEEE/CVF Conference on Computer Vision and Pattern Recognition}, pp.\  26417--26427, 2024.

\bibitem[Reid et~al.(2024)Reid, Savinov, Teplyashin, Lepikhin, Lillicrap, Alayrac, Soricut, Lazaridou, Firat, Schrittwieser, et~al.]{reid2024gemini}
Reid, M., Savinov, N., Teplyashin, D., Lepikhin, D., Lillicrap, T., Alayrac, J.-b., Soricut, R., Lazaridou, A., Firat, O., Schrittwieser, J., et~al.
\newblock Gemini 1.5: Unlocking multimodal understanding across millions of tokens of context.
\newblock \emph{arXiv preprint arXiv:2403.05530}, 2024.

\bibitem[Reimers \& Gurevych(2019)Reimers and Gurevych]{reimers2019sentence}
Reimers, N. and Gurevych, I.
\newblock Sentence-bert: Sentence embeddings using siamese bert-networks.
\newblock \emph{arXiv preprint arXiv:1908.10084}, 2019.

\bibitem[Robinson et~al.(2020)Robinson, Chuang, Sra, and Jegelka]{robinson2020contrastive}
Robinson, J., Chuang, C.-Y., Sra, S., and Jegelka, S.
\newblock Contrastive learning with hard negative samples.
\newblock \emph{arXiv preprint arXiv:2010.04592}, 2020.

\bibitem[Rozenberszki et~al.(2022)Rozenberszki, Litany, and Dai]{rozenberszki2022language}
Rozenberszki, D., Litany, O., and Dai, A.
\newblock Language-grounded indoor 3d semantic segmentation in the wild.
\newblock In \emph{European Conference on Computer Vision}, pp.\  125--141. Springer, 2022.

\bibitem[Sarkar et~al.(2024)Sarkar, Ebrahimi, Etemad, Beirami, Ar{\i}k, and Pfister]{sarkar2024mitigating}
Sarkar, P., Ebrahimi, S., Etemad, A., Beirami, A., Ar{\i}k, S.~{\"O}., and Pfister, T.
\newblock Mitigating object hallucination via data augmented contrastive tuning.
\newblock \emph{arXiv preprint arXiv:2405.18654}, 2024.

\bibitem[Sharkawy \& Koustoumpardis(2022)Sharkawy and Koustoumpardis]{sharkawy2022human}
Sharkawy, A.-N. and Koustoumpardis, P.~N.
\newblock Human--robot interaction: A review and analysis on variable admittance control, safety, and perspectives.
\newblock \emph{Machines}, 10\penalty0 (7):\penalty0 591, 2022.

\bibitem[Team et~al.(2024)Team, Mesnard, Hardin, Dadashi, Bhupatiraju, Pathak, Sifre, Rivi{\`e}re, Kale, Love, et~al.]{team2024gemma}
Team, G., Mesnard, T., Hardin, C., Dadashi, R., Bhupatiraju, S., Pathak, S., Sifre, L., Rivi{\`e}re, M., Kale, M.~S., Love, J., et~al.
\newblock Gemma: Open models based on gemini research and technology.
\newblock \emph{arXiv preprint arXiv:2403.08295}, 2024.

\bibitem[Touvron et~al.(2023)Touvron, Martin, Stone, Albert, Almahairi, Babaei, Bashlykov, Batra, Bhargava, Bhosale, et~al.]{touvron2023llama}
Touvron, H., Martin, L., Stone, K., Albert, P., Almahairi, A., Babaei, Y., Bashlykov, N., Batra, S., Bhargava, P., Bhosale, S., et~al.
\newblock Llama 2: Open foundation and fine-tuned chat models.
\newblock \emph{arXiv preprint arXiv:2307.09288}, 2023.

\bibitem[Vaswani et~al.(2017)Vaswani, Shazeer, Parmar, Uszkoreit, Jones, Gomez, Kaiser, and Polosukhin]{vaswani2017attention}
Vaswani, A., Shazeer, N., Parmar, N., Uszkoreit, J., Jones, L., Gomez, A.~N., Kaiser, {\L}., and Polosukhin, I.
\newblock Attention is all you need.
\newblock \emph{Advances in neural information processing systems}, 30, 2017.

\bibitem[Vedantam et~al.(2015)Vedantam, Lawrence~Zitnick, and Parikh]{vedantam2015cider}
Vedantam, R., Lawrence~Zitnick, C., and Parikh, D.
\newblock Cider: Consensus-based image description evaluation.
\newblock In \emph{Proceedings of the IEEE conference on computer vision and pattern recognition}, pp.\  4566--4575, 2015.

\bibitem[Wang et~al.(2023)Wang, Huang, Zhao, Zhang, and Zhao]{wang2023chat3d}
Wang, Z., Huang, H., Zhao, Y., Zhang, Z., and Zhao, Z.
\newblock Chat-3d: Data-efficiently tuning large language model for universal dialogue of 3d scenes.
\newblock \emph{arXiv preprint arXiv:2308.08769}, 2023.

\bibitem[Xu et~al.(2023)Xu, Wang, Wang, Chen, Pang, and Lin]{xu2023pointllm}
Xu, R., Wang, X., Wang, T., Chen, Y., Pang, J., and Lin, D.
\newblock Pointllm: Empowering large language models to understand point clouds.
\newblock \emph{arXiv preprint arXiv:2308.16911}, 2023.

\bibitem[Yan et~al.(2024)Yan, Wang, Huang, Zhou, Yin, Galstyan, Yin, and Chen]{yan2024contrastive}
Yan, T., Wang, F., Huang, J.~Y., Zhou, W., Yin, F., Galstyan, A., Yin, W., and Chen, M.
\newblock Contrastive instruction tuning.
\newblock \emph{arXiv preprint arXiv:2402.11138}, 2024.

\bibitem[Yang et~al.(2024)Yang, Chen, Madaan, Iyengar, Qian, Fouhey, and Chai]{yang20243d-grand}
Yang, J., Chen, X., Madaan, N., Iyengar, M., Qian, S., Fouhey, D.~F., and Chai, J.
\newblock 3d-grand: A million-scale dataset for 3d-llms with better grounding and less hallucination.
\newblock \emph{arXiv preprint arXiv:2406.05132}, 2024.

\bibitem[Yu et~al.(2022)Yu, Tang, Rao, Huang, Zhou, and Lu]{yu2022point}
Yu, X., Tang, L., Rao, Y., Huang, T., Zhou, J., and Lu, J.
\newblock Point-bert: Pre-training 3d point cloud transformers with masked point modeling.
\newblock In \emph{Proceedings of the IEEE/CVF conference on computer vision and pattern recognition}, pp.\  19313--19322, 2022.

\bibitem[Zhang et~al.(2024)Zhang, Wen, Fu, Wang, Zhang, Wang, and Jin]{zhang2024llava-hd}
Zhang, Y.-F., Wen, Q., Fu, C., Wang, X., Zhang, Z., Wang, L., and Jin, R.
\newblock Beyond llava-hd: Diving into high-resolution large multimodal models.
\newblock \emph{arXiv preprint arXiv:2406.08487}, 2024.

\bibitem[Zheng et~al.(2023)Zheng, Ke, Zhang, and Huang]{zheng2023click}
Zheng, C., Ke, P., Zhang, Z., and Huang, M.
\newblock Click: Controllable text generation with sequence likelihood contrastive learning.
\newblock In \emph{Findings of the Association for Computational Linguistics: ACL 2023}, pp.\  1022--1040, 2023.

\bibitem[Zhou et~al.(2024{\natexlab{a}})Zhou, Liu, Xu, Iyer, Sun, Mao, Ma, Efrat, Yu, Yu, et~al.]{zhou2024lima}
Zhou, C., Liu, P., Xu, P., Iyer, S., Sun, J., Mao, Y., Ma, X., Efrat, A., Yu, P., Yu, L., et~al.
\newblock Lima: Less is more for alignment.
\newblock \emph{Advances in Neural Information Processing Systems}, 36, 2024{\natexlab{a}}.

\bibitem[Zhou et~al.(2024{\natexlab{b}})Zhou, Wang, Ma, Liu, Huang, and Wang]{zhou2023uni3d}
Zhou, J., Wang, J., Ma, B., Liu, Y.-S., Huang, T., and Wang, X.
\newblock Uni3d: Exploring unified 3d representation at scale.
\newblock In \emph{International Conference on Learning Representations (ICLR)}, 2024{\natexlab{b}}.

\bibitem[Zhu et~al.(2023)Zhu, Ma, Chen, Deng, Huang, and Li]{zhu20233d-vista}
Zhu, Z., Ma, X., Chen, Y., Deng, Z., Huang, S., and Li, Q.
\newblock 3d-vista: Pre-trained transformer for 3d vision and text alignment.
\newblock In \emph{Proceedings of the IEEE/CVF International Conference on Computer Vision}, pp.\  2911--2921, 2023.

\end{thebibliography}
\bibliographystyle{icml2025}

\newpage
\appendix
\definecolor{mygreen}{HTML}{3cb44b}

\newcommand{\forinline}{ \textcolor{magenta!90!black} }
\newcommand{\assign}{\leftarrow}
\newcommand{\var}{\texttt}
\newcommand{\FuncCall}[2]{\texttt{\bfseries #1(#2)}}
\newcommand{\VarSty}[1]{\textnormal{\ttfamily\color{blue!90!black}#1}\unskip}
\newcommand{\PredSty}[1]{\textnormal{\ttfamily\color{mygreen!90!black}#1}\unskip}

\onecolumn
\section{Appendix}
\subsection{Architecture}
CL3DOR is a generative model designed to produce language responses by integrating both point cloud data and textual inputs. The primary components of CL3DOR are the 3D scene encoder, the point-language connector, and a pre-trained LLM.

\paragraph{3D Scene Encoder}
The 3D scene encoder comprises a pre-trained point cloud encoder and a spatial transformer. In our experiments, we utilize PointBERT~\cite{yu2022point}, a Transformer-based model~\cite{vaswani2017attention} pre-trained on the ShapeNet dataset~\cite{chang2015shapenet}, which contains over 50,000 3D models. Each model in ShapeNet is represented by 8,192 point clouds, a resolution we maintain in CL3DOR to ensure high fidelity in point cloud data. This approach allows us to fully leverage PointBERT’s robust feature embedding capabilities. Importantly, PointBERT remains frozen throughout all three training stages of CL3DOR, meaning its weights are not updated, thus preserving its pre-trained knowledge.

The spatial transformer~\cite{chen2022language}, widely used in 3D Large Multimodal Models (3D LMMs)~\cite{zhu20233d-vista,huang2023chat3dv2,huang2023embodied}, is crucial for learning spatial relationships between multiple objects within a 3D scene. This module explicitly calculates pairwise spatial relations and integrates them with standard self-attention to enhance spatial reasoning capabilities. We employ a three-layer spatial transformer with 8 heads to process the object-centric features produced by PointBERT, ultimately generating object tokens for the LLM. For other configurations, we maintain the default settings as outlined in~\cite{chen2022language}.

\paragraph{Point-Language Connector}
To align the output vector space of the 3D scene encoder with the input vector space of the large language model, we use projection layers consisting of two layers activated by the GeLU~\cite{hendrycks2016gaussian}. The parameters of this module are updated throughout all training stages.

\paragraph{Large Language Model}
We select a decoder-only Transformer-based Large Language Model (LLM) as our backbone, specifically LLaMA3-8B-Instruct\footnote{\url{https://github.com/meta-llama/llama3}}~\cite{dubey2024llama}. The LLM's input includes the system message, 3D scene point clouds, and language instructions. Before being fed into the LLM, the 3D scene point clouds are transformed into special tokens.

To seamlessly integrate the 3D point cloud data, we expand the LLM's vocabulary with three special tokens: \texttt{<point\_start>}, \texttt{<point\_patch>}, and \texttt{<point\_end>}, and train their corresponding embeddings. The \texttt{<point\_patch>} token represents each object in the 3D scene, so if a 3D scene comprises \(N\) objects, there will be \(N\) \texttt{<point\_patch>} tokens. The \texttt{<point\_start>} token is placed at the beginning of the point cloud sequence, and the \texttt{<point\_end>} token is appended at the end, encapsulating the \texttt{<point\_patch>} tokens within the LLM's input sequence.

\subsection{Implementation details}

CL3DOR is trained through a three-stage paradigm, with the hyperparameters for each stage consolidated into Table~\ref{merged_hyperparameters_columnwise}. It provides a comprehensive overview of the hyperparameters used for 3D-object alignment tuning, spatial alignment tuning, and spatial contrastive instruction tuning, facilitating direct comparison across the stages. Due to limited available resources, all evaluations are conducted using a single inference run, rather than multiple runs to generate statistical values.

\begin{table}[htbp]
\centering
\caption{Hyperparameters for Different Tuning Tasks}
\label{merged_hyperparameters_columnwise}
\begin{tabular}{lccc}
\toprule
\textbf{Hyperparameter} & \textbf{3D-Object Alignment} & \textbf{Spatial Alignment} & \textbf{Spatial Contrastive Instruction} \\
\midrule
Optimizer & AdamW & AdamW & AdamW \\
Weight decay & 5e-2 & 5e-2 & 5e-2 \\
Betas & [0.9, 0.999] & [0.9, 0.999] & [0.9, 0.999] \\
Learning rate & 2e-3 & 1e-4 & 5e-6 \\
Type of GPUs & NVIDIA A6000 & NVIDIA A6000 & NVIDIA A6000 \\
Number of GPUs & 4 & 4 & 4 \\
Accumulate gradient batches & 2 & 2 & 2 \\
Batch size per GPU (total) & 64 (256) & 16 (64) & 16 (64) \\
Max steps & 2.5K & 4.8K & 1.6K \\
\bottomrule
\end{tabular}
\end{table}




\subsection{Details in Dataset}
\label{sec:reference_examples}
\paragraph{Coordinate Alignment} When training with ReferIt3D, Scan2Cap, or SQA3D, the positions of objects and instructions are incorporated. Therefore, it is necessary to align the coordinates of the captions and the 3D scene within a unified world coordinate system. To achieve this, we translate the origin of each scene to the mean of the points that compose the scene, thereby aligning it with the actual intended position.

\paragraph{Instruction in SQA3D} In SQA3D, the captions include both the instructions (e.g., "Which direction should I go if I want to throw litter?") and the instructor's current situation (e.g., "I am sitting on the chair with a bag behind me."). During training, the situation and the instruction are combined into a single instruction.

\paragraph{System Prompt} Below is the system prompt used for all training stages of CL3DOR. This prompt is prepended to the instruction as input to the model.
\begin{mdframed}
\emph{You are able to understand the visual content that the user provides and assist the user with a variety of tasks using natural language. You should follow the instructions carefully.} 
\end{mdframed}

\paragraph{Response Format Prompting} Recent studies~\cite{liu2024improved, cha2024honeybee} in the domain of 2D Large Multimodal Models have demonstrated the effectiveness of format prompts for eliciting responses in a desired structure. Building on these findings, format prompts are applied in Stage 3 and tailored to each specific task: \textit{Answer the question using a single word or phrase} for 3D Question Answering, \textit{Describe it briefly} for 3D-Object-in-the-Scene Captioning, and \textit{Please answer with yes or no} for 3D-Object Existence.

Additionally, the instructions used in each task are randomly selected from the same instruction candidates as those in~\cite{huang2023embodied}.

\paragraph{3D Scene Captioning Data Generation}
We use GPT-4o to generate detailed descriptions of 3D scenes. Although GPT-4o is designed for 2D images, it can understand and describe the spatial aspects of 3D scenes based on 2D images captured from various angles. To achieve results aligned with our intentions, we carefully design prompts that effectively guide GPT-4o. As shown in Table~\ref{tab:prompt_3Dscenecaption}, these prompts detail the features of the input data and provide few-shot examples to clarify the desired output. Additionally, we find that a single top-view image is insufficient for generating a comprehensive 3D scene description. Therefore, in addition to the top-view image, we capture four additional images from different angles, all oriented toward the center of the 3D scene.

\paragraph{Dataset Examples}
We provide additional examples of datasets used during each training stage of CL3DOR, specifically for stages 2 and 3, where 3D scenes are used as input. In stage 2, which focuses on spatial alignment tuning, and stage 3, which involves spatial contrastive instruction tuning, corresponding dataset examples can be found in Tables~\ref{dataset_example1}, \ref{dataset_example2}, and \ref{dataset_example3}.

\paragraph{Hard Negative Response Generation}
To enable effective contrastive learning in CL3DOR, we use triplet data consisting of a question, a positive response, and a hard negative response. These hard negatives are plausible but incorrect, helping the model learn fine-grained distinctions. Our pipeline augments 3D-scene datasets with these hard negatives generated by GPT-4o, excluding the 3D object existence task. In this section, we introduce the prompts and details used to generate hard negatives for the 3D Question Answering and 3D-Object-in-the-Scene Captioning datasets.

The prompt used for generating 3D Question Answering data with GPT-4o is shown in Table~\ref{tab:prompt_3DQA}. The input context includes a top-view 2D image, scene metadata, the original question, and the ground truth answer. The scene metadata provides a list of objects present in the input scene. The input prompt is also designed to filter out illogical QA pairs, thereby improving the overall quality of the training data.

The prompt used for generating 3D-object-in-the-scene captions with GPT-4o is presented in Table~\ref{tab:prompt_3Dobjectinthescene}. To create hard negatives, we introduce hallucinations related to the object's characteristics, intentionally making the prompt noisy. The specified object and the corresponding positive answer are included in the prompt. Since the descriptions are detailed enough to generate plausible negative answers based purely on text, the focus is on maximizing the generation of hard negatives using textual descriptions alone, without requiring visual input.

\subsection{Additional Experiments}
\label{sec:appendix_ablation}

\paragraph{Effect of Odds Ratio Loss}
In this section, we compare different loss functions to examine the importance of the OR term in the objective function and explore various hyperparameters for $\lambda$, as specified in Eq.~\eqref{eq:combined_loss}. For the main experiment, we set the $\lambda$ as 3e-1.

As discussed in the Discussion section, we conducted an experiment using the Probability Ratio (PR) instead of the OR term in the objective function to assess the significance of the OR term. Following the methodology outlined in \cite{hong2024reference}, we applied this approach to datasets related to 3D scene understanding and reasoning, specifically ScanQA, SQA3D, and Scan2Cap, as well as the 3D hallucination dataset 3D-POPE, which were used in the main experiment. We tested various $\lambda$ values of 5e-2, 1e-1, 3e-1, and 5e-1 with the OR term. It is important to note that when $\lambda$ is set to 0, the objective function reduces to the standard SFT used in other language modeling tasks.

As shown in Tables \ref{tab:appendix_scanqa}, \ref{tab:appendix_sqa3d}, \ref{tab:appendix_scan2cap}, and \ref{tab:appendix_pope}, the results using the OR term outperformed those using PR and SFT across all metrics for ScanQA, SQA3D, and Scan2Cap, regardless of the $\lambda$ value, demonstrating the robustness of our method. For the 3D-POPE dataset, the results with the OR term were consistently higher than those with SFT and were comparable to or slightly better than those with PR. These findings indicate that the OR term is highly effective for 3D scene-related tasks. Additionally, the generally higher performance of PR compared to SFT suggests that the hard negatives we generated are effective and that applying contrastive learning to 3D LMMs significantly enhances performance.

\paragraph{Qualitative Results}
We conduct additional qualitative evaluations of CL3DOR on ScanQA, SQA3D, and Scan2Cap datasets. As shown in Tables~\ref{qualitative_ex1}, \ref{qualitative_ex2}, and \ref{qualitative_ex3}, analyzing the responses generated by CL3DOR to various instructions shows that the model effectively understands and reasons about spatial relationships.

\begin{table}[htbp]
\centering
\caption{An ablation study of hyperparameter $\lambda$ on ScanQA benchmark. Values in parentheses represent refined exact-match scores.}
\label{tab:appendix_scanqa}
\small
\begin{tabularx}{\textwidth}{>{\centering\arraybackslash}c>{\centering\arraybackslash}X>{\centering\arraybackslash}X>{\centering\arraybackslash}X>{\centering\arraybackslash}X>{\centering\arraybackslash}X}
\toprule
\multicolumn{6}{c}{\textbf{\large{ScanQA (val)}}} \\
\midrule
\textbf{$\lambda$} & CIDEr & BLEU-4 & METEOR & ROUGE-L & EM@1 (refined) \\
\midrule
0 (SFT) & 99.9 & 14.0 & 19.1 & 49.9 & 26.1 (54.3) \\
5e-2 & 109.8 & 15.7 & 20.9 & 52.6 & 26.1 (52.7) \\
1e-1 & 110.1 & 16.0 & 20.9 & 52.6 & 26.1 (52.8) \\
3e-1 & 110.4 & 16.7 & 21.0 & 52.5 & 25.8 (52.9) \\
5e-1 & 108.0 & 15.6 & 20.8 & 51.8 & 24.9 (51.5) \\
\midrule
PR & 106.1 & 14.9 & 20.5 & 51.5 & 25.4 (51.5) \\
\bottomrule
\end{tabularx}
\end{table}


\begin{table}[htbp]
\caption{An ablation study of hyperparameter $\lambda$ on SQA3D benchmark. Values in parentheses represent refined exact-match scores.}
\label{tab:appendix_sqa3d}
\centering
\begin{tabularx}{\textwidth}{>{\centering\arraybackslash}c>{\centering\arraybackslash}X>{\centering\arraybackslash}X}
\toprule
\multicolumn{3}{c}{\textbf{\large{SQA3D (test)}}} \\
\midrule
\textbf{$\lambda$} & \textbf{EM@1} & \textbf{EM@1 (refined)} \\
\midrule
0 (SFT) & 45.92 & 48.3 \\
5e-2 & 50.0 & 52.7 \\
1e-1 & 50.0 & 52.9 \\
3e-1 & 51.6 & 54.4 \\
5e-1 & 49.7 & 52.3 \\
\midrule
PR & 50.0 & 52.7 \\
\bottomrule
\end{tabularx}
\end{table}



\begin{table}[htbp]
\caption{An ablation study of hyperparameter $\lambda$ on Scan2Cap benchmark.}
\label{tab:appendix_scan2cap}
\centering
\begin{tabularx}{\textwidth}{>{\centering\arraybackslash}c>{\centering\arraybackslash}X>{\centering\arraybackslash}X>{\centering\arraybackslash}X>{\centering\arraybackslash}X>{\centering\arraybackslash}X}
\toprule
\multicolumn{6}{c}{\textbf{\large{Scan2Cap (val)}}} \\
\midrule
\textbf{$\lambda$} & \textbf{CIDEr} & \textbf{BLEU-4} & \textbf{METEOR} & \textbf{ROUGE-L} & \textbf{Sim} \\
\midrule
0 (SFT) & 91.4 & 36.5 & 27.6 & 60.5 & 66.3 \\
5e-2 & 94.6 & 36.3 & 27.6 & 60.4 & 67.2 \\
1e-1 & 94.5 & 36.1 & 27.6 & 60.2 & 67.2 \\
3e-1 & 93.4 & 36.0 & 27.6 & 60.1 & 67.1 \\
5e-1 & 93.1 & 36.1 & 27.6 & 60.2 & 67.1 \\
\midrule
PR & 91.2 & 35.7 & 27.5 & 60.1 & 66.7 \\
\bottomrule
\end{tabularx}
\end{table}


\begin{table}[htbp]
\caption{An ablation study of hyperparameter $\lambda$ on 3D-POPE benchmark.}
\label{tab:appendix_pope}
\centering
\small 
\begin{tabularx}{\textwidth}{>{\raggedright\arraybackslash}l c>{\centering\arraybackslash}X>{\centering\arraybackslash}X>{\centering\arraybackslash}X>{\centering\arraybackslash}X>{\centering\arraybackslash}X}
\toprule
\multicolumn{7}{c}{\textbf{\large{3D-POPE}}} \\
\midrule
\textbf{Type} & \textbf{$\lambda$} & \textbf{Precision} & \textbf{Recall} & \textbf{F1} & \textbf{Accuracy} & \textbf{Yes (\%)} \\
\midrule
\multirow{6}{*}{\makecell{\textit{\textbf{Random}}}} 
& 0 (SFT) & 63.3 & 74.5 & 68.4 & 65.6 & 58.8 \\
& 5e-2 & 72.1 & 76.9 & 74.4 & 73.6 & 53.3 \\
& 1e-1 & 72.6 & 77.4 & 74.9 & 74.1 & 53.3 \\
& 3e-1 & 74.0 & 80.0 & 75.9 & 76.9 & 54.1 \\
& 5e-1 & 75.4 & 81.9 & 78.5 & 77.6 & 54.4 \\
\cmidrule(lr{0.1em}){2-7}
& PR & 73.0 & 77.7 & 75.3 & 74.5 & 53.2 \\
\midrule
\multirow{6}{*}{\makecell{\textit{\textbf{Popular}}}} 
& 0 (SFT) & 57.7 & 74.5 & 65.0 & 59.9 & 64.5 \\
& 5e-2 & 60.5 & 76.9 & 67.7 & 63.3 & 63.6 \\
& 1e-1 & 60.8 & 77.4 & 68.1 & 63.7 & 63.7 \\
& 3e-1 & 61.5 & 80.0 & 65.0 & 69.6 & 65.0 \\
& 5e-1 & 61.3 & 81.9 & 70.2 & 65.2 & 66.8 \\
\cmidrule(lr{0.1em}){2-7}
& PR & 62.1 & 77.7 & 69.0 & 65.1 & 62.6 \\
\midrule
\multirow{6}{*}{\makecell{\textit{\textbf{Adversarial}}}} 
& 0 (SFT) & 56.3 & 74.4 & 64.1 & 58.4 & 66.0 \\
& 5e-2 & 58.1 & 77.3 & 67.7 & 63.3 & 63.6 \\
& 1e-1 & 58.5 & 78.3 & 68.1 & 63.7 & 63.7 \\
& 3e-1 & 58.9 & 80.7 & 68.1 & 62.2 & 68.5 \\
& 5e-1 & 58.7 & 82.7 & 70.2 & 65.2 & 66.8 \\
\cmidrule(lr{0.1em}){2-7}
& PR & 60.2 & 78.5 & 68.1 & 63.2 & 65.3 \\
\bottomrule
\end{tabularx}
\end{table}

\begin{table}[ht!]\centering
\caption{The prompt provided to GPT-4o for generating 3D Scene Captioning is utilized in Spatial Alignment Tuning. It incorporates five images of a room captured from different angles, including a top-view, to articulate the overall ambiance of the room and the spatial relationships between objects within the scene.}
\begin{minipage}{\columnwidth}\vspace{0mm} \centering
    \begin{tcolorbox}[width=\columnwidth]
        \small
        \begin{tabular}{p{\columnwidth}}
        \VarSty{\textbf{Prompt for 3D Scene Captioning:}} \\

    Your task is to provide a detailed description of a space-based solely on provided multi-view images, without mentioning the image itself. Describe the space from different perspectives to give a comprehensive understanding of its layout, features, and atmosphere. \\
    \\
    Your descriptions must comply with the following constraints:\\
    - Describe each detail as comprehensively as possible.\\
    - Avoid mentioning external information, such as the perspective of the image.\\
    - Format your description into a single paragraph without any special symbols.\\
    - Include detailed descriptions of each object and its color within the scene.\\
    - You do provide captions based on visuals, but in your responses, never use the word 'images.' Instead, use different expressions to write the descriptions.\\
    - In some images, certain objects or scenes may not be clearly visible and there may be ambiguous parts. In such cases, do not attempt to forcefully add captions based on arbitrary guesses.\\
\\
\\
    For example 1)\\
    Description :
    In this depiction, the scene unfolds within a room richly adorned with contemporary furnishings marked by vibrant splashes of red and understated neutral tones. Predominantly, a plush red sofa commands attention, complemented by a matching red ottoman centered on a large gray area rug. This lounge area is flanked by a variety of shelving units filled with an assortment of items that suggest a living space-cum-studio. The bookshelves are packed with knick-knacks and books, indicating a personal and used space. Visible soft gray pillows tastefully arranged on the sofa add a touch of comfort and style, suitable for relaxation and guest entertainment. Nearby, a modernistic white floor lamp provides functional yet aesthetic lighting, and small, colorful art pieces on the walls contribute a creative and lively ambiance to the space. Other details, such as a white decorative vase and green cabinetry, enhance the eclectic and personalized feel of the room, suggesting this is a well-lived and cherished space.
    \\
    \\
    For example 2)\\
    Description :
    In this detailed scene, we are presented with a diverse visual compilation of a living area, kitchen, and utility spaces. The living area features a comfortable couch, alongside which stands a floor lamp offering illumination. Adjacent to this is a compact yet stocked kitchenette showcasing essential appliances like a microwave and a coffee maker, both indicative of modern living comforts. A dining corner is discerned, identified by a wooden table surrounded by chairs, which indicates a common area for meals. Notable domestic features such as a washing machine and a basket suggest routine laundry activities. The ambiance is complemented by personal touches evident from pictures and posters adorning the walls, adding a layer of individuality and lived-in appeal. The floor, visible as a continuous surface stretching across the area, bears no significant wear, enhancing the overall aesthetics. The inclusion of a bathroom, though partially visible, with standard fittings underscores the comprehensive functionality of the layout. Instruments such as a guitar imply leisure and creative pursuits within this space, enriching the scene's narrative of daily life and recreation. The presence of storage solutions, seen as bookshelves brimming with books and sundry items, ensures that the essentials are neatly organized, promoting an orderly environment. Collectively, these elements piece together a vivid portrayal of a dwelling designed for comfort, convenience, and personal expression.
    \\
    \\
    Description : 
\\

        \end{tabular}
    \end{tcolorbox}
\vspace{-2mm}
\label{tab:prompt_3Dscenecaption}
\end{minipage}
\end{table}

\begin{table}[ht!]\centering
\caption{The prompt given to GPT-4o for generating hard negative samples for 3D Question Answering used in Spatial Contrastive Instruction Tuning. In this prompt, \VarSty{\{question\}} represents the question from the existing dataset, and \VarSty{\{chosen\_value\}} is the corresponding answer used as the positive response. \VarSty{\{object\_dict\}} denotes the list of objects present in the scene, which, along with the top-view image, is used to generate the hard negatives. Upon reviewing the existing data, we identify some instances where the questions and answers are unreasonable; thus, an additional prompt is included to filter out these cases.}
\begin{minipage}{\columnwidth}\vspace{0mm} \centering
    \begin{tcolorbox}[width=\columnwidth]
        \small
        \begin{tabular}{p{\columnwidth}}
        \VarSty{\textbf{Prompt for 3D Question Answering:}} \\

        Generate an ‘incorrect answer' for a given question about a 3D scene. The ‘incorrect answer' should be a hard negative: plausible yet wrong, challenging to distinguish from the ground truth answer. The input context includes a top-view 2D image, scene metadata, the original question, and the ground truth answer. \\
        \\
        Conditions for generating the ‘incorrect answer': \\
        1. The ‘incorrect answer' should demonstrate one or more of the following perceptual blindness issues: \\
        1-1. Failure to recognize the existence of objects that are actually present in the scene. \\
        1-2. Misidentification of object locations within the scene. For ambiguous direction questions, generate an ‘incorrect answer' with a clearly incorrect object. (e.g., Q: “Where is the beige wooden working table placed?" GT: “right of tall cabinet" Incorrect answer: “left of nightstand") \\
        1-3. Misattribution of object attributes such as size, shape, or color. \\
        \\
        2. Create a short, concise ‘incorrect answer' that: \\
        2-1. Is composed of objects listed in the ‘all objects' information. \\
        2-2. Follows the same format as the GT. \\
        2-3. Is short and concise. Do not provide any other type of response like ‘-', ‘output:', or ‘incorrect answer:'. \\
        \\
        3. For certain problematic cases, generate one of the following responses instead of an ‘incorrect answer': \\
        3-1. “REMOVE THIS SAMPLE: Insufficient image context" - If the image alone does not provide enough context to answer the question or determine the GT answer. \\
        3-2. “REMOVE THIS SAMPLE: Unreasonable GT" - If the GT is unreasonable (e.g., Q: “What is under the table?" GT: “Yes"). \\
        3-3. “REMOVE THIS SAMPLE: Directional ambiguity" - If the question or GT involves ambiguous directions such as left or right. \\
        \\
        Input context: \\
        - Question: \VarSty{\{question\}} \\
        - Ground Truth (GT): \VarSty{\{chosen\_value\}} \\
        - All objects in the scene: \VarSty{\{object\_dict\}} \\
        \end{tabular}
    \end{tcolorbox}
\vspace{-2mm}
\label{tab:prompt_3DQA}
\end{minipage}
\end{table}

\begin{table}[ht!]\centering
\caption{The prompt provided to GPT-4o for generating hard negative samples for a 3D-Object-in-the-Scene in Spatial Contrastive Instruction Tuning. \VarSty{\{Target\_object\}} refers to a specific object within the scene, and \VarSty{\{Description\}} represents the spatial relationships between this object and other objects in the scene in the positive response. 
No visual image is included in this prompt.}
\begin{minipage}{\columnwidth}\vspace{0mm} \centering
    \begin{tcolorbox}[width=\columnwidth]
        \small
        \begin{tabular}{p{\columnwidth}}
        \VarSty{\textbf{Prompt for 3D-Object-in-the-Scene Captioning:}} \\

    Below is a description of a specific object and its positional relationship to another object. Based on this description, generate a plausible but slightly noisy description by adding some hallucinations, such as incorrect colors or incorrect positional relationships. The new description should not deviate significantly from the original form and should maintain a similar structure to the following set. Even if the resulting description contains grammatical errors, ensure it retains the original format as much as possible:
\\
\\
Ex1) a black tv, in the direction from the entrance and from the outside, will be on the right side of the blue curtain. on the left of the tv is a small bike.\\
Ex2) there is a white toilet. placed in the corner of the bath.\\
Ex3) there is a white plastic clothes handler. placed next to the bin in the corner.
\\
\\
Target Object: \VarSty{\{Target\_Object\}} \\
Description:  \VarSty{\{Description\}}
\\
\\
Generate a new description:
        \end{tabular}
    \end{tcolorbox}
\vspace{-2mm}
\label{tab:prompt_3Dobjectinthescene}
\end{minipage}
\end{table}

\begin{table}[ht!]\centering
\caption{Examples of training datasets for Stage 2 and Stage 3 using \emph{Scene\_0005} from the ScanNet dataset.}
\begin{minipage}{\columnwidth}\vspace{0mm} \centering
    \begin{tcolorbox}[width=\columnwidth]
        \footnotesize
        \begin{tabular}{p{\columnwidth}}
        \VarSty{\textbf{3D Scene}} \\

        \includegraphics[width=0.8\columnwidth]{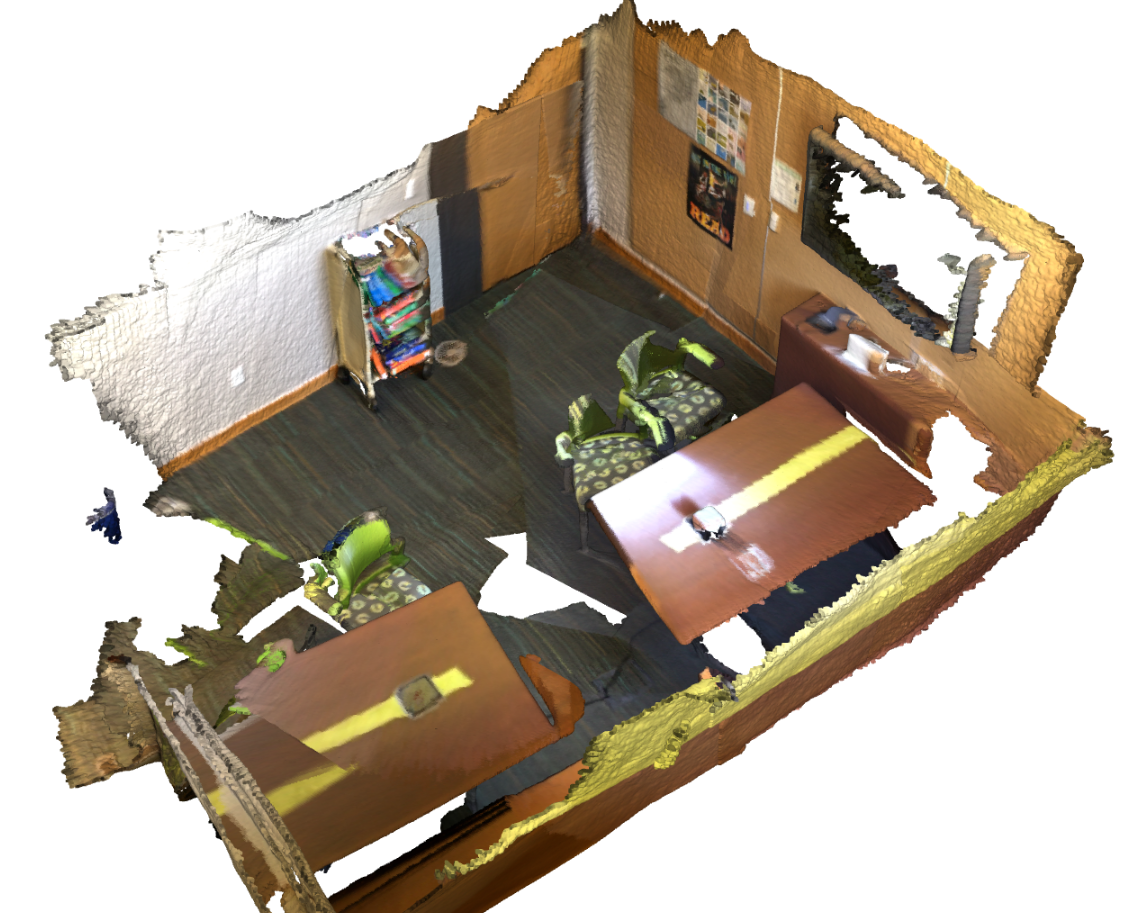} \\
        \vspace{2mm}

        \VarSty{\textbf{[Stage 2] 3D Object-in-the-Scene Captioning}} \\
\textbf{Instruction} : I'd like a straightforward description of the object at the 3D location you mentioned. \\
\textbf{Response} : Table that is not near the tv \\

\VarSty{\textbf{[Stage 2] 3D Scene Captioning}} \\
\textbf{Instruction} : Could you elaborate extensively on what this scene contains? \\
\textbf{Response} : In this setting, we encounter a cozy and welcoming library-like space, designed to inspire relaxation and study. The room features several large tables, each bathed in the glow from above by modern, circular light fixtures. These tables are surrounded by comfortable armchairs upholstered in a vivid pattern of green, yellow, and black, suggesting a lively yet sophisticated decor scheme. The flooring, a rich, dark wood, adds warmth to the room and contrasts sharply with the brightness of the tables and chairs. Along one wall, a collection of colorful posters and bulletin boards, brimming with papers and notices, injects a personal and vibrant touch, illustrating the space’s communal and informative role. Nearby, a tall bookshelf filled to the brim with books spans an entire wall, reinforcing the academic function of the area. Additionally, a small nook houses a multi-colored cart packed with materials, further emphasizing the room's purpose for study and leisure reading. The area is enclosed by walls painted in a soothing cream tone, interrupted only by windows and doorways that break up the expanse with promises of other rooms beyond. This arrangement presents an inviting scene where community and learning intersect comfortably.\\

\VarSty{\textbf{[Stage 3] 3D Object-in-the-Scene Captioning}} \\
\textbf{Instruction} : Focusing on the 3D scene, describe the object at the indicated spot. Describe it briefly. \\
\textbf{Positive} : a wooden green chair with arms. it is located very near the wall and close to the table. \\
\textbf{Negative} : a metal blue chair with arms. it is located very near the window and close to the sofa. \\

\VarSty{\textbf{[Stage 3] 3D Question Answering}} \\
\textbf{Instruction} : What type of table is on the right side of the room? Answer the question using a single word or phrase. \\
\textbf{Positive} : wooden \\
\textbf{Negative} : laminated \\


\VarSty{\textbf{[Stage 3] 3D-Object Existence}} \\
\textbf{Instruction} : Does the room contain any refrigerator? Please answer with yes or no. \\
\textbf{Positive} : no \\
\textbf{Negative} : yes \\

        \end{tabular}
    \end{tcolorbox}
\vspace{-2mm}
\label{dataset_example1}
\end{minipage}
\end{table}

\begin{table}[ht!]\centering
\caption{Examples of training datasets for Stage 2 and Stage 3 using \emph{Scene\_0017} from the ScanNet dataset.}
\begin{minipage}{\columnwidth}\vspace{0mm} \centering
    \begin{tcolorbox}[width=\columnwidth]
        \small
        \begin{tabular}{p{\columnwidth}}
        \VarSty{\textbf{3D Scene}} \\
        \includegraphics[width=0.6\columnwidth]{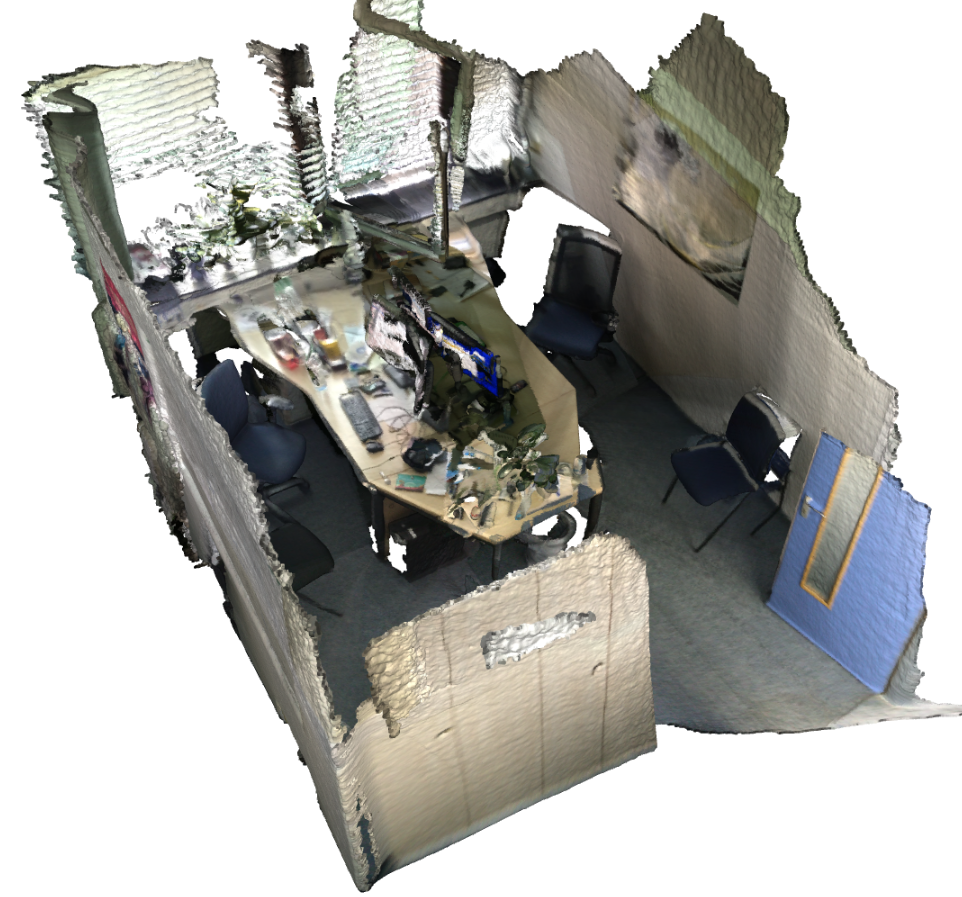} \\
        \vspace{2mm}
        
        \VarSty{\textbf{[Stage 2] 3D Object-in-the-Scene Captioning}} \\
\textbf{Instruction} : Describe the primary features of the object at the indicated 3D spot. \\
\textbf{Response} : The toilet paper hanging on the wall by the toilet \\

\VarSty{\textbf{[Stage 2] 3D Scene Captioning}} \\
\textbf{Instruction} : Could you provide an in-depth description of this scene? \\
\textbf{Response} : The space indicates a functionally arranged office environment, with various elements suggesting a bustling, creative, and tech-savvy setting. At the office's core sits a large rectangular desk cluttered with numerous computer monitors, suggesting a collaborative working atmosphere where technology plays a central role. The desk is surrounded by rotatable office chairs covered in a deep blue fabric, and it hosts a myriad of gadgets, cables, and personal items hinting at a space utilized for substantial digital work and project development. The floor is carpeted in a subdued gray, offering a neutral backdrop that contrasts with the vibrantly colored technological equipment and personal items scattered across the desk, such as a bright red coffee cup and various note-filled papers that reflect the spontaneous and dynamic nature of work done here. Adjacent to the work area, a side table holds a couple of printers and copious papers, emphasizing the utility and function-driven aspect of the room. Walls in lighter shades, possibly beige or pale gray, carry detailed illustrations and planning charts, bordered by various personal photos and decorations that add a personal touch to an otherwise highly professional setup. A pale blue door on the corridor offers a cleanly defined egress, complementing the orderly, albeit busy, character of this workspace. The ambience reflects a balance between high-paced professional engagements and a personable, relaxed creative flair, ideal for innovative pursuits and focused productivity.\\

\VarSty{\textbf{[Stage 3] 3D Object-in-the-Scene Captioning}} \\
\textbf{Instruction} : I'd like a straightforward description of the object at the 3D location you mentioned. Describe it briefly. \\
\textbf{Positive} : the trash can is located under the desk on the outside end. it is black and round and has a liner. \\
\textbf{Negative} : the trash can is located under the table on the inside end. it is blue and square and has a lid. \\

\VarSty{\textbf{[Stage 3] 3D Question Answering}} \\
\textbf{Instruction} : What is underneath an office desk? Answer the question using a single word or phrase. \\
\textbf{Positive} : trash can \\
\textbf{Negative} : plant \\

\VarSty{\textbf{[Stage 3] 3D-Object Existence}} \\
\textbf{Instruction} : Have you noticed any desk in the room? Please answer with yes or no. \\
\textbf{Positive} : yes \\
\textbf{Negative} : no \\
        \end{tabular}
    \end{tcolorbox}
\vspace{-2mm}
\label{dataset_example2}
\end{minipage}
\end{table}

\begin{table}[ht!]\centering
\caption{Examples of training datasets for Stage 2 and Stage 3 using \emph{Scene\_0238} from the ScanNet dataset.}
\begin{minipage}{\columnwidth}\vspace{0mm} \centering
    \begin{tcolorbox}[width=\columnwidth]
        \small
        \begin{tabular}{p{\columnwidth}}
        \VarSty{\textbf{3D Scene}} \\
        \includegraphics[width=0.8\columnwidth]{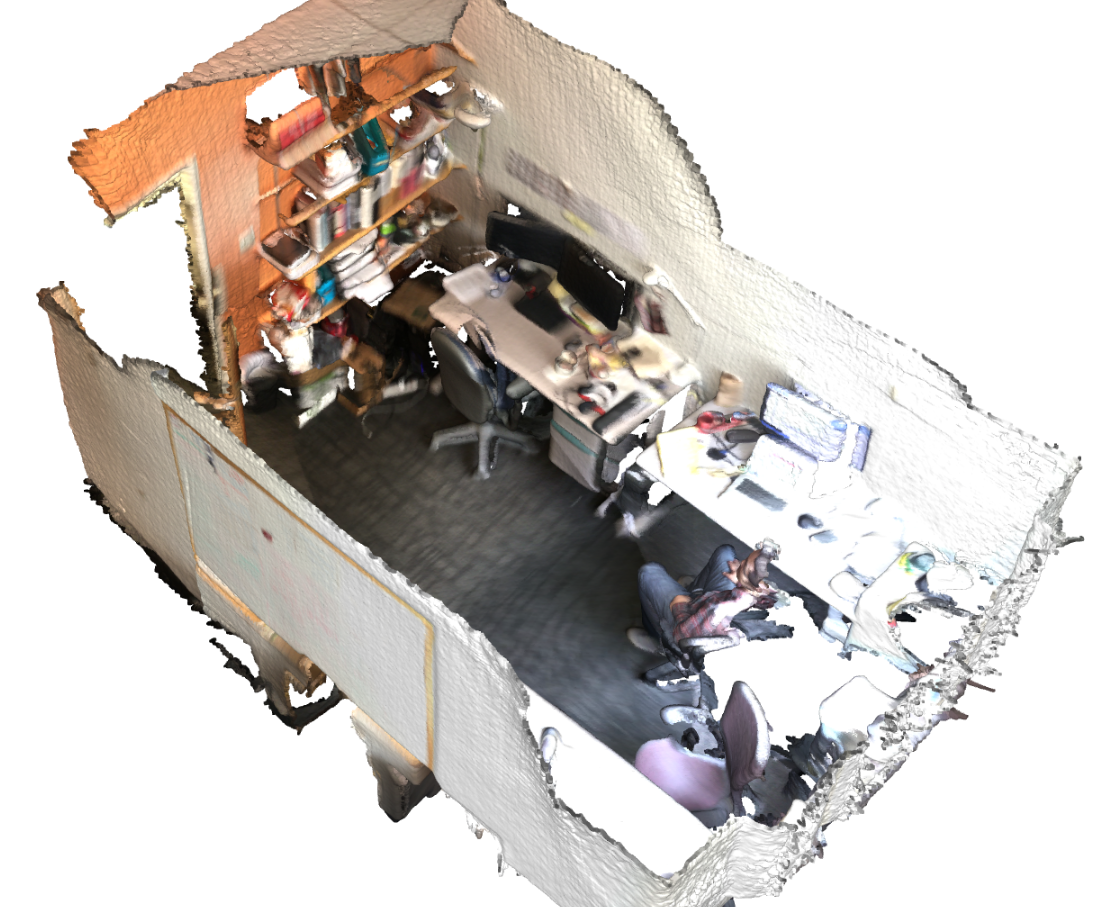} \\
        \vspace{2mm}
        
        \VarSty{\textbf{[Stage 2] 3D Object-in-the-Scene Captioning}} \\
\textbf{Instruction} : Identify the object located at the specific point in the 3D environment. \\
\textbf{Response} : The large box closest to the open door.\\

\VarSty{\textbf{[Stage 2] 3D Scene Captioning}} \\
\textbf{Instruction} : In a few words, how would you depict this scene? \\
\textbf{Response} : In this compact and bustling artist's workshop, the main area is predominantly occupied by a large desk that hosts an array of artistic and electronic equipment, suggesting a hybrid space of traditional and digital creation. The desk is cluttered with items like sketchbooks, loose papers, various art supplies, and a laptop, all under the glow of a desk lamp, adding a functional yet chaotic charm to the area. Surrounding the desk, the walls are densely populated with shelves burdened with an eclectic mix of further art supplies, books, and personal knick-knacks, appearing to provide both inspiration and organization to the inhabitant. The presence of a whiteboard filled with notes and sketches adjacent to the desk serves as a storyboard or a planner, pointing towards ongoing projects or ideas. The floor, covered in a dark carpet, contrasts sharply with the lighter colored walls and is littered casually with more items, possibly overflow from the creative undertakings. Additional seating options like a small chair and a wheeled office chair suggest a space that accommodates long hours of work and the occasional visitor. The overall atmosphere is one of creative fervor, marked by a casual disarray that often accompanies artistic processes.\\

\VarSty{\textbf{[Stage 3] 3D Object-in-the-Scene Captioning}} \\
\textbf{Instruction} : Focus on the object at the given 3D location and describe it. Describe it briefly. \\
\textbf{Positive} : the chair is the northeastern-most one in the room. the chair is purple and has five legs. \\
\textbf{Negative} : the chair is the southwestern-most one in the room. the chair is green and has four legs. \\

\VarSty{\textbf{[Stage 3] 3D Question Answering}} \\
\textbf{Instruction} : How many black office chairs are there? Answer the question using a single word or phrase. \\
\textbf{Positive} : 4 \\
\textbf{Negative} : 3 \\


\VarSty{\textbf{[Stage 3] 3D-Object Existence}} \\
\textbf{Instruction} : Have you noticed any recycling bin in the room? Please answer with yes or no. \\
\textbf{Positive} : no \\
\textbf{Negative} : yes \\
        \end{tabular}
    \end{tcolorbox}
\vspace{-2mm}
\label{dataset_example3}
\end{minipage}
\end{table}

\begin{table}[ht!]\centering
\caption{Qualitative results for \emph{Scene\_0050} in the ScanNet dataset.}
\begin{minipage}{\columnwidth}\vspace{0mm} \centering
    \begin{tcolorbox}[width=\columnwidth]
        \small
        \begin{tabular}{p{\columnwidth}}
        \VarSty{\textbf{3D Scene}} \\
        \includegraphics[width=0.9\columnwidth]{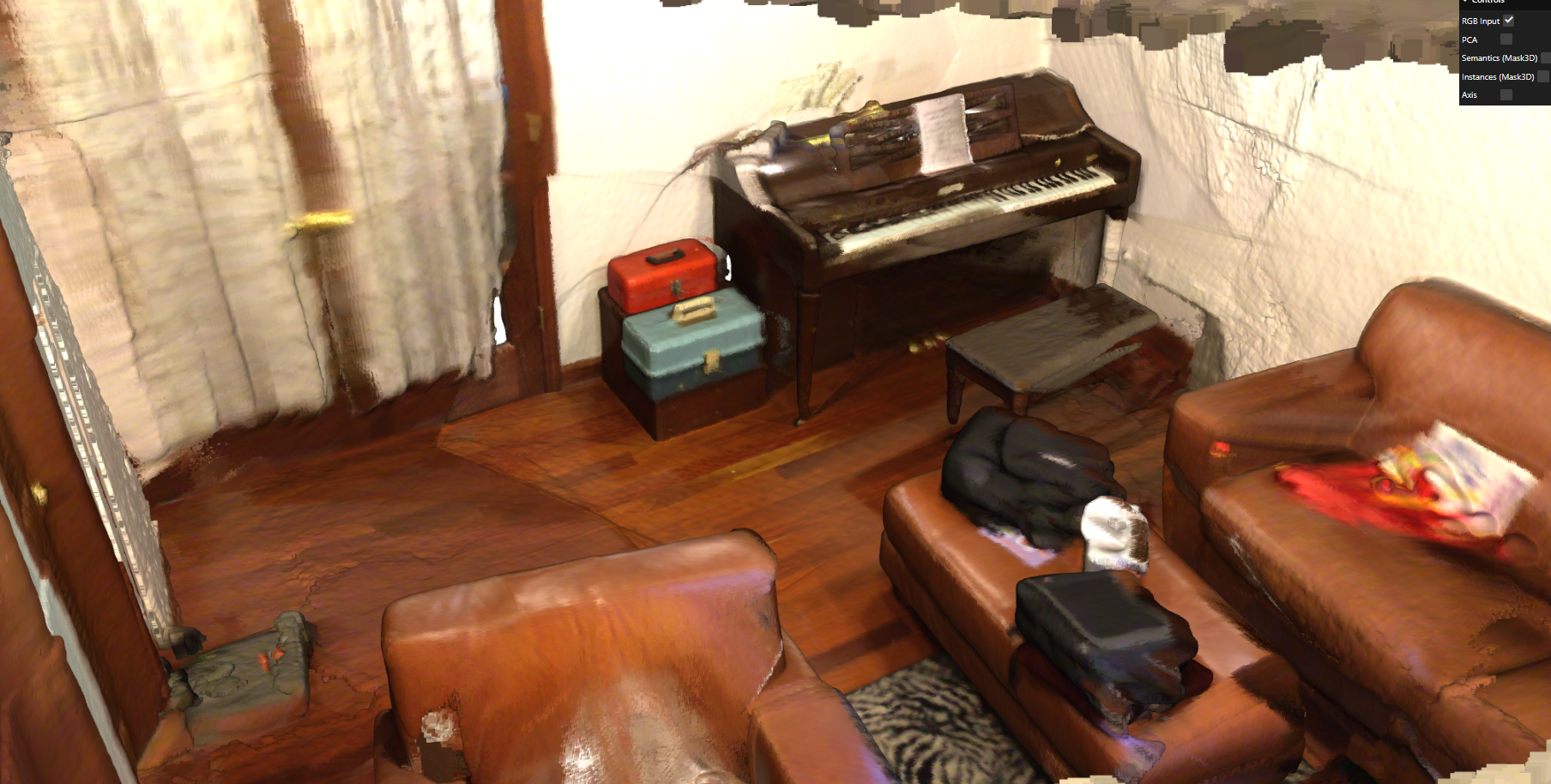} \\
        \vspace{2mm}
        
        \VarSty{\textbf{3D Object-in-the-Scene Captioning - Scan2Cap}} \\
\textbf{Instruction} : What was the object like at the chosen location in the 3D scene? Describe it briefly. \\
\textbf{Ground truth} : this is a brown piano bench. it is in front of the piano. \\
\textbf{CL3DOR} : this is a brown wooden piano bench. it is in front of a piano. there is a blue  tool box and a red tool box next to the piano that the bench is in front of . \\

\VarSty{\textbf{3D Question Answering - ScanQA}} \\
\textbf{Instruction} : What is a box of on the ottoman? Answer the question using a single word or phrase. \\
\textbf{Ground truth} : tissue, white box of tissues \\
\textbf{CL3DOR} : tissue box \\

\VarSty{\textbf{3D Question Answering - SQA3D}} \\
\textbf{Instruction} : Are the curtains open or closed to my left? \\
\textbf{Ground truth} : closed \\
\textbf{CL3DOR} : closed \\
        \end{tabular}
    \end{tcolorbox}
\vspace{-2mm}
\label{qualitative_ex1}
\end{minipage}
\end{table}










\begin{table}[ht!]\centering
\caption{Qualitative results for \emph{Scene\_0084} in the ScanNet dataset.}
\begin{minipage}{\columnwidth}\vspace{0mm} \centering
    \begin{tcolorbox}[width=\columnwidth]
        \small
        \begin{tabular}{p{\columnwidth}}
        \VarSty{\textbf{3D Scene}} \\
        \includegraphics[width=0.9\columnwidth]{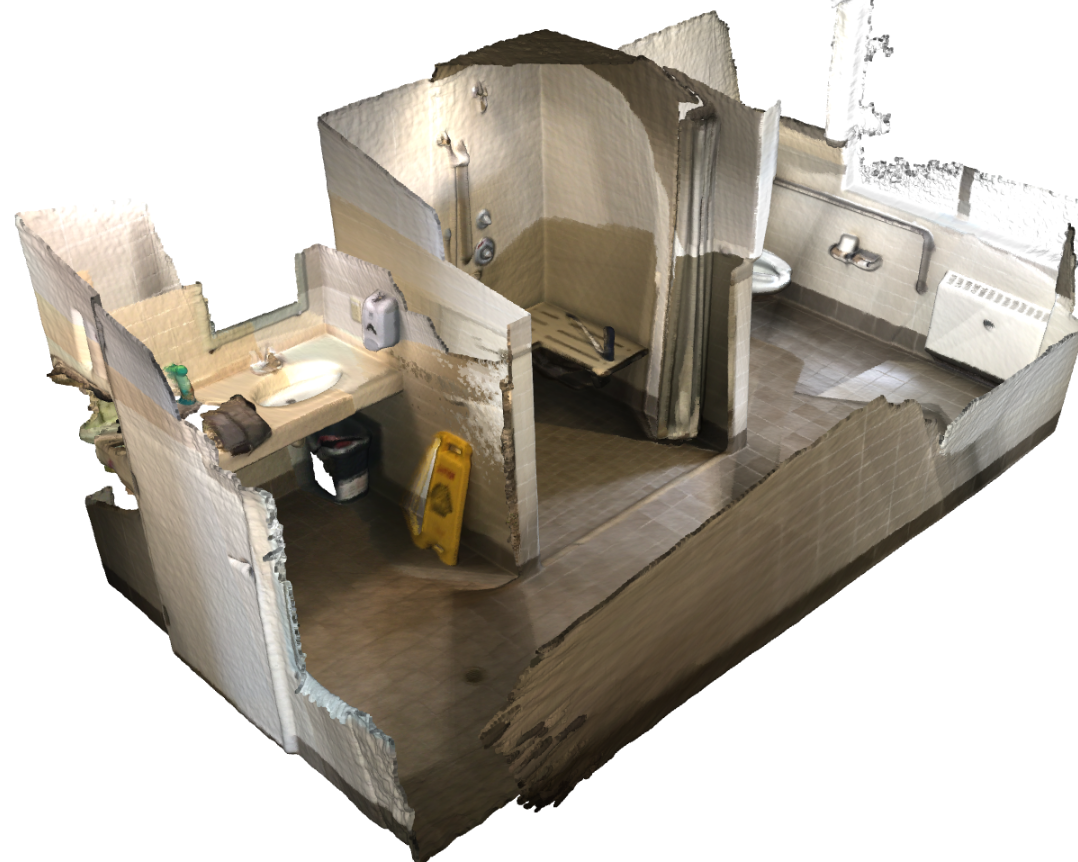} \\
        \vspace{2mm}
        
        \VarSty{\textbf{3D Object-in-the-Scene Captioning - Scan2Cap}} \\
\textbf{Instruction} : Say what the object looks like at the picked location in the 3D space. Describe it briefly. \\
\textbf{Ground truth} : it is a tan bathroom counter. the counter is next to the door . \\
\textbf{CL3DOR} : there is a rectangular bathroom counter. it is over trash cans. \\

\VarSty{\textbf{3D Question Answering - ScanQA}} \\
\textbf{Instruction} : What color soap dispenser is above the sink? Answer the question using a single word or phrase. \\
\textbf{Ground truth} : white, grey \\
\textbf{CL3DOR} : white \\

\VarSty{\textbf{3D Question Answering - SQA3D}} \\
\textbf{Instruction} : To exit can I just walk through the door or I have to open the door? \\
\textbf{Ground truth} : open \\
\textbf{CL3DOR} : open \\

        \end{tabular}
    \end{tcolorbox}
\vspace{-2mm}

\label{qualitative_ex2}
\end{minipage}
\end{table}










\begin{table}[ht!]\centering
\caption{Qualitative results for \emph{Scene\_0221} in the ScanNet dataset.}
\begin{minipage}{\columnwidth}\vspace{0mm} \centering
    \begin{tcolorbox}[width=\columnwidth]
        \small
        \begin{tabular}{p{\columnwidth}}
        \VarSty{\textbf{3D Scene}} \\
        \includegraphics[width=0.9\columnwidth]{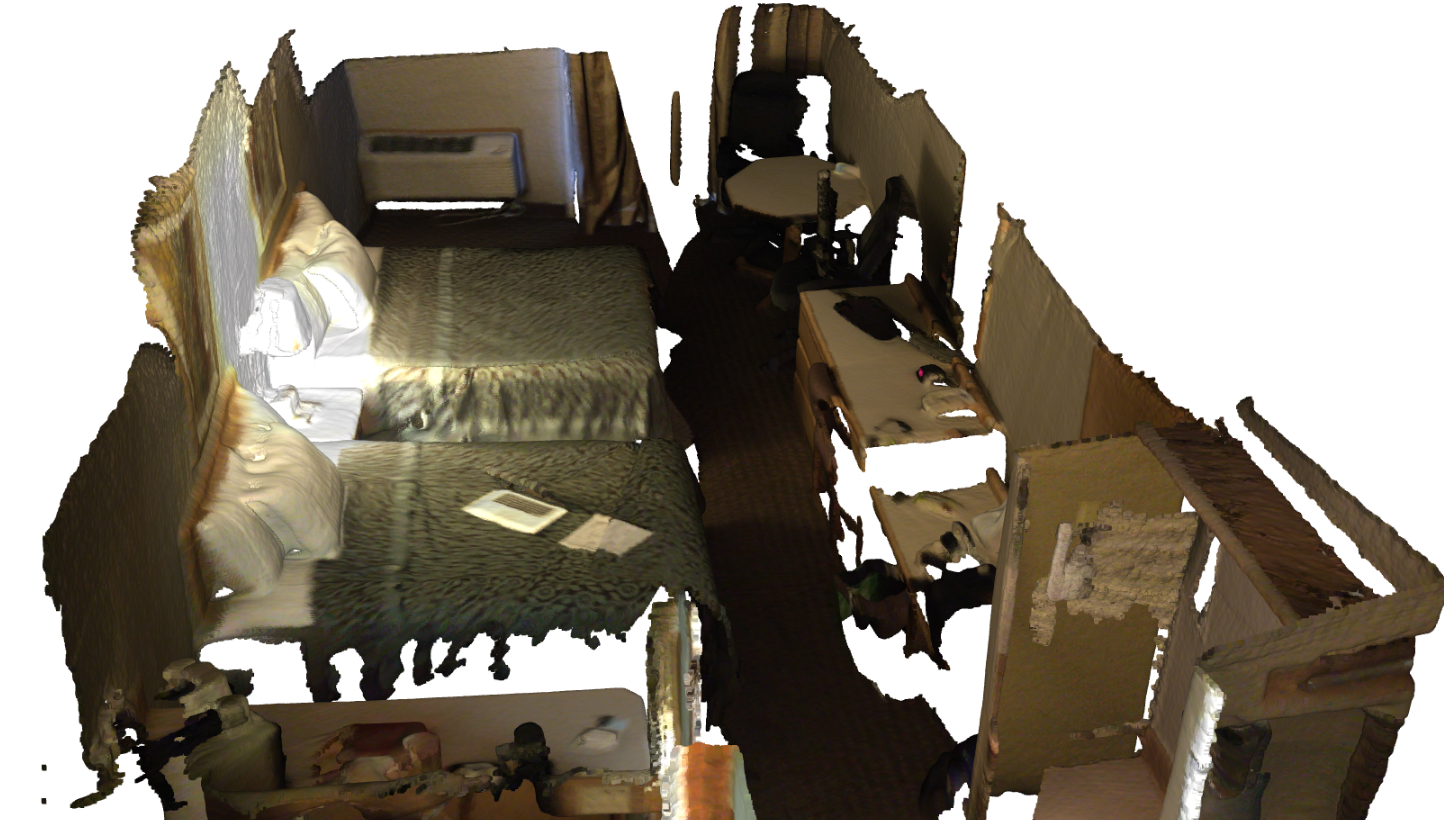} \\
        \vspace{2mm}
        
        \VarSty{\textbf{3D Object-in-the-Scene Captioning - Scan2Cap}} \\
\textbf{Instruction} : Can you clarify the object at the picked spot in the 3D scene? Describe it briefly. \\
\textbf{Ground truth} :  this is a black rolling desk chair . it is located along the wall to the left of the dresser . \\
\textbf{CL3DOR} : this is a black chair. it is at a round table. \\

\VarSty{\textbf{3D Question Answering - ScanQA}} \\
\textbf{Instruction} : Where is the white pillow placed? Answer the question using a single word or phrase. \\
\textbf{Ground truth} : next to another same pillow, near headboard of both beds \\
\textbf{CL3DOR} : on bed \\

\VarSty{\textbf{3D Question Answering - SQA3D}} \\
\textbf{Instruction} : What color are the pillows on the bed to my left? \\
\textbf{Ground truth} : white \\
\textbf{CL3DOR} : white \\

        \end{tabular}
    \end{tcolorbox}
\vspace{-2mm}

\label{qualitative_ex3}
\end{minipage}
\end{table}











\end{document}